\documentclass{article}

\usepackage{microtype}
\usepackage{graphicx}
\usepackage{subcaption}
\usepackage{booktabs} 
\usepackage[utf8]{inputenc}
\usepackage{amsmath}
\usepackage{amssymb} 
\usepackage{mathtools}
\usepackage{wrapfig}
\usepackage[table]{xcolor}

\usepackage{url}
\usepackage[dvipsnames]{xcolor}       
\usepackage{algorithm}

\usepackage{wrapfig}
\usepackage{amsthm}
\usepackage[most]{tcolorbox} 
\usepackage{lipsum} 
\usepackage{booktabs,multirow}

\usepackage{booktabs,multirow,siunitx}
\usepackage{threeparttable}

\usepackage{listings}    

\definecolor{codeblue}{rgb}{0.8,0.2,0.2}
\definecolor{codegray}{rgb}{0.5,0.5,0.5}
\definecolor{highlight}{RGB}{220,235,255}

\definecolor{stdgray}{gray}{0.45}

\definecolor{LightBlue}{RGB}{80,160,255}

\PassOptionsToPackage{
  colorlinks=true,
  citecolor=LightBlue,
  linkcolor=Orange,
  urlcolor=Orange
}{hyperref}

\usepackage[preprint]{icml2026}

\usepackage{hyperref}
\hypersetup{
  colorlinks=true,
  citecolor=LightBlue,
  linkcolor=Orange,
  urlcolor=Orange
}

\usepackage[capitalize,noabbrev]{cleveref}

\usepackage{amsmath}
\usepackage{amssymb}
\usepackage{mathtools}
\usepackage{amsthm}

\usepackage[capitalize,noabbrev]{cleveref}

\theoremstyle{plain}
\newtheorem{theorem}{Theorem}[section]
\newtheorem{proposition}[theorem]{Proposition}
\newtheorem{lemma}[theorem]{Lemma}

\theoremstyle{definition}

\theoremstyle{remark}

\usepackage[textsize=tiny]{todonotes}

\icmltitlerunning{{Y-Shaped Generative Flows}}

\begin{document}

\twocolumn[
  \icmltitle{Y-Shaped Generative Flows}




  \begin{icmlauthorlist}
    \icmlauthor{Arip Asadulaev}{yyy}
    \icmlauthor{Semyon Semenov}{yyy}
    \icmlauthor{Abduragim Shtanchaev}{yyy}
    \icmlauthor{Eric Moulines}{yyy}
    \icmlauthor{Fakhri Karray}{yyy}
    \icmlauthor{Martin Takac}{yyy}
  \end{icmlauthorlist}

  \icmlaffiliation{yyy}{MBZUAI}

  \icmlcorrespondingauthor{Arip Asadulaev}{arip.asadulaev@mbzuai.ac.ae}

  \icmlkeywords{ICML, Generative Models, Y-shaped, Flow models}

  \vskip 0.3in
]

\printAffiliationsAndNotice{}  

\begin{abstract}
Modern continuous-time generative models typically induce \emph{V-shaped} flows: each sample travels independently along a nearly straight trajectory from the prior to the data. Although effective, this independent movement overlooks the hierarchical structures that exist in real-world data. To address this, we introduce \emph{Y-shaped generative flows}, a framework in which samples travel together along shared pathways before branching off to target-specific endpoints. Our formulation is theoretically justified, yet remains practical, requiring only minimal modifications to standard velocity-driven models. We implement this through a scalable, neural network-based training objective. Experiments on synthetic, image, and biological datasets demonstrate that our method recovers hierarchy-aware structures, improves distributional metrics over strong flow-based baselines, and reaches targets in fewer steps.
\end{abstract}

\vspace{-2mm}
\section{Introduction}
\label{sec:intro}
Recent advances in generative modeling are heavily based on continuous-time flows. In this framework, a model learns a velocity field to transport the mass from a simple starting distribution to a complex target distribution~\citep{ho2020denoising, lipman2022flow}. Current state-of-the-art methods, such as flow matching, treat every data point as an isolated traveler moving along its own independent path~\citep{tong2023improving, liu2022flow}. For example, moving a starting point $x_0$ to two different targets, $y_1$ and $y_2$, requires two completely separate trajectories that do not interact. We refer to this uncoordinated pattern as a \emph{V-shaped flow}, as the objective does not provide a mechanism for the paths to merge.

\begin{figure}[t!]
    \centering
    \includegraphics[width=0.48\textwidth]{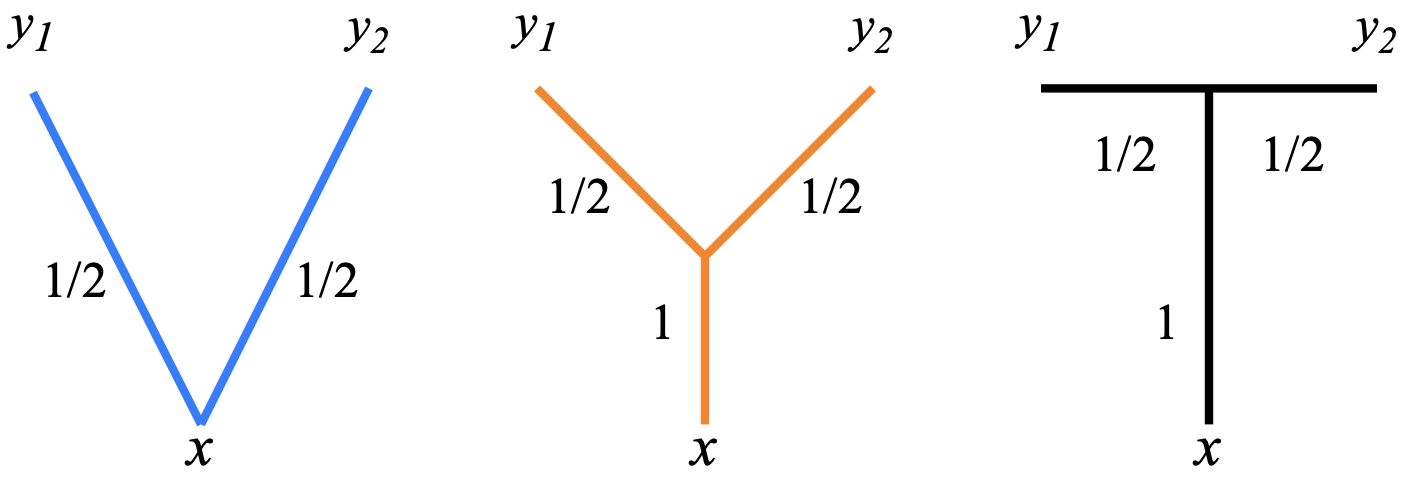}
    \caption{\emph{Blueprint of flow shapes}. Conceptually in a V-flow, the mass separates and moves from source $x$ along straight lines. In Y and T flows, the mass moves together and then splits into targets.}
    \vspace{-6mm}
    \label{fig:Y-shaped_toy}
\end{figure}

However, we ask: \emph{is a V-shaped flow always the most natural way to translate between real-world data distributions?} We argue that it often is not. Real-world data usually possess a hierarchical structure, groupings, and taxonomies that V-flows ignore. In applied mathematics, a closely related question was raised in \underline{branched} optimal transport theory. This theory is inspired by natural and engineered systems designed for efficiency, such as vascular systems, trees, river basins, and urban supply networks \citep{buttazzo2003optimal, xia2003optimal, bernot2005traffic}. In these systems, it is more efficient to transport a large mass together along a main \emph{trunk} before splitting it, rather than sending many small units individually. Translating this branching principle into generative modeling suggests that trajectories should be allowed to merge toward well-represented modes and then branch out to reach diverse targets. In contrast to conventional independent V-flows, we introduce \emph{Y-shaped generative flows} (Figure \ref{fig:Y-shaped_toy}). 

This structural shift also addresses a second major challenge in current generative modeling research: efficiency. Many recent V-shaped approaches attempt to reduce the number of steps required to generate a sample using techniques such as mean flows~\citep{geng2025mean}, shortcut models~\citep{frans2024one}, and \textit{jumps}~\citep{holderrieth2024generator}. We argue that a Y-shaped framework is a more fundamental solution for faster generation. In a Y-flow, samples targeting central modes of a distribution share a common path and reach their destination quickly, while more unique or peripheral samples branch off later. This allows the model to naturally allocate steps based on complexity, rather than forcing a one-size-fits-all trajectory.

\textbf{Contribution.} We formally introduce \emph{Y-shaped generative flows}, a continuous-time model that encourages shared transport before branching. Our method is based on a \emph{concave velocity-power} action. This objective was not previously studied, and we theoretically show that our method actually promotes the concentration of trajectories followed by branching. To implement this, we develop a scalable, neural network-based training objective. This objective balances the time-integrated branched action with a target reaching the boundary to ensure the flow matches the target data. Empirically, we demonstrate that our models learn branched trajectories on synthetic, image, and single-cell datasets, improving performance metrics (e.g., W$_1$/W$_2$/MMD) over strong flow-based baselines.

\vspace{-2mm}
\section{Background}
\label{sec:premilinaries}
\textbf{Notation.}
Let $\Omega \subset \mathbb{R}^d$ be a compact and convex domain. We denote by $\mathcal{P}_2(\Omega)$ the space of probability measures on $\Omega$ with a finite second moment. When a measure $\mu$ admits a density relative to the Lebesgue measure, we denote its density by $\rho$ (i.e., $d\mu(x) = \rho(x)dx$). Vectors are column vectors, $|\cdot|$ is the standard Euclidean norm, $\nabla\!\cdot$ is the divergence operator, and  $|\cdot|_F$ is the Frobenius norm. The notation $d\mathcal{H}^1$ represents the differential element with respect to the \textit{Hausdorff measure} in one dimension. In practical terms, for a curve or any 1-dimensional set in space, the measure $\mathcal{H}^1$ simply calculates its length. 

\textbf{Continuous Normalizing Flows (CNFs)} proposed by \cite{chen2018neural} are generative models that define a probability density path through a neural ordinary differential equation (ODE) to morph one probability distribution, $\mu_0$, into another, $\mu_1$: $\frac{d}{dt} x_t = v_\theta(x_t,t)$, where $x_{t=0} \sim \mu_0.$
Let $\Phi_t$ be the flow map associated with this ODE, which transports a particle from its initial condition at time $0$ to its location at time $t$. The pushforward density $\rho_t = (\Phi_t)_\# \rho_0$ evolving under this dynamics \emph{necessarily} satisfies \emph{continuity equation} with the parameterized velocity field $v_\theta$:
\begin{align}
\label{eq:continuity}
  \partial_t \rho_t + \nabla \cdot (\rho_t v_t) &= 0
  \quad \text{on } \Omega \times (0,1), \\ 
  \rho_{t=0}&=\rho_0,\quad \rho_{t=1}=\rho_1.\nonumber
\end{align}
A key result is the instantaneous change of variables formula, which describes how the log density evolves along a trajectory: $\frac{d}{dt}\,\log \rho_t(x_t) = -\,\nabla\!\cdot v_\theta(x_t,t).$ This allows for a likelihood calculation by integrating this quantity over time. Training can be done by directly maximizing likelihood.

\textbf{Monge--Kantorovich Optimal Transport (OT)}
\label{sec:static_ot}
seeks a way to morph one probability distribution with minimal effort, as quantified by a cost function $c(x, y)$. The Monge formulation seeks a deterministic map $T: \Omega \to \Omega$ that pushes $\mu_0$ to $\mu_1$ and minimizes the total cost $\int c(x, T(x)) d\mu_0(x)$ \citep{villani2008optimal}. This problem can be ill-posed, its relaxation, the Kantorovich problem, searches over \emph{couplings} (joint distributions) $\pi \in \Pi(\mu_0,\mu_1)$ with marginals $\mu_0$ and $\mu_1$: $\inf_{\pi \in \Pi(\mu_0,\mu_1)}
  \int_{\Omega \times \Omega} c(x,y)\, d\pi(x,y).$
For cost $c(x,y)=\|x-y\|^2$, the square root of the solution is the Wasserstein-2 distance. 

\textbf{Benamou--Brenier OT} 
frames transportation as a continuous-time problem. For flow $\Phi_t$, we have a pushforward density $\rho_t = (\Phi_t)_\# \rho_0$ evolving under the velocity field $v$. The Benamou--Brenier theorem states that the squared Wasserstein distance equals the minimal \emph{kinetic energy} of such a flow:
\begin{equation}
  \label{eq:bb}
  W_2^2(\mu_0,\mu_1)
  =
  \inf_{\rho_t,\, v_t}
  \int_0^1 \!\! \int_{\Omega} |v_t(x)|^2 \,\rho_t(x)\, dx\, dt \quad \text{s.t. Eq. \eqref{eq:continuity}} .
\end{equation}
The minimizers of this problem are constant-speed geodesics in the Wasserstein space. When an optimal transport map $T$ exists for the static problem (e.g., when $\mu_0$ is absolutely continuous), the geodesic is given by \emph{interpolation}: $\rho_t = ((1-t)\mathrm{Id}+tT)_\# \rho_0$. The corresponding velocity field is constant along the trajectories, $v_t(x_t) = T(x_0)-x_0$.

\section{On Shapes of Flows}
\label{subsec:branched_ot_bb}
\emph{What is the most natural way to control the shape of a generative flow?}
To approach this question, we draw inspiration from the mathematical domain that studies the organization of branching structures in nature and human-made systems, such as rivers, blood vessels, and traffic networks \citep{xia2003optimal, brasco2011benamou}. The core of all branched methods is the concave transport cost. Typically of the form $m^{\alpha} \ell$, where $m$ is the mass and $\ell$ is the distance, and $\alpha \in (0, 1)$ is a branching coefficient. Since $(m_1 + m_2)^{\alpha} < m_1^{\alpha} + m_2^{\alpha}$ for $\alpha<1$, this cost structure incentivizes the masses to travel together along shared paths. Transporting a combined mass along a single path is cheaper than transporting a mass $m$ along a few separate parallel paths~\citep{santambrogio2015optimal}.  


In the branched transport formulation introduced by~\cite{xia2003optimal}, the transport system is modeled as a flow network represented by $F$. Physically, $F$ acts as the momentum or \emph{mass flux}. It encodes both the direction of movement and the amount of mass flowing at any given point. Transport is concentrated along a geometric network of 1-dimensional paths, denoted by $\Gamma$. In each segment of the path, let $\theta$ represent the amount (or thickness) of mass flowing. Mass conservation is strictly enforced by the continuity equation $\nabla \cdot F = \mu_0 - \mu_1$. Here, the divergence $\nabla \cdot F$ acts as a local accounting system, ensuring that the net outflow from any point matches the difference between the source distribution ($\mu_0$) and the target distribution ($\mu_1$). The goal is to minimize the transport cost, $E_\alpha(F)$, defined as:
\begin{equation}
    E_\alpha(F) = \int_{\Gamma} \theta^{\alpha} \, d\mathcal{H}^1
\end{equation}
Unlike simple distance minimization, this integral calculates the mass cost ($\theta^\alpha$) multiplied by the distance traveled ($d\mathcal{H}^1$). Because $\theta^\alpha$ is sub-additive, it is \emph{cheaper} to group the mass into thick trunks rather than sending the particles along separate, thin paths. However, because this cost function is highly non-convex and defined over a complex space of singular fields, directly minimizing this functional is computationally intractable.

Another perspective on branched optimal transport is the Benamou–Brenier formulation studied by~\cite{brasco2011benamou}. The idea is to consider the  transport problem \eqref{eq:bb}, but with a \emph{concave} cost function. With $\alpha \in (0,1)$ being the branching parameter, we define the cost that measures the \textit{work} of moving a mass $m$ over a distance $\ell$. Identifying the length $\ell$ with the velocity $v$ and the mass $m$ with the time-dependent probability density $\rho_t$, the dynamic branched transport is the following:
\begin{equation}
  \label{eq:branched_bb}
  B^{\alpha}(\mu_0,\mu_1)
  \;=\;
  \inf_{\substack{\rho,\,v}}
  \int_{0}^{1}
  \sum_{i\in I_t} |v_{t,i}|\,(\rho_{t,i})^{\alpha}
  \,dt, \quad\text{s.t. Eq. } \eqref{eq:continuity}.
\end{equation}
In this formulation, the cost \eqref{eq:branched_bb} should be finite, the measure $\rho(t,\cdot)$ \underline{must be purely atomic} for (almost every) time $t$. This means that the mass is concentrated at a countable set of points $\{x_{t,i}\}_{i\in I_t}$:
$
    \rho(t,\cdot) = \sum_{i\in I_t} \rho_{t,i}\,\delta_{x_{t,i}},  \text{where } \rho_{t,i} = \rho(t, \{x_{t,i}\}).
$
The associated momentum is then $F = \rho v = \sum_{i\in I_t} v_{t,i}\,\rho_{t,i}\,\delta_{x_{t,i}}$, where $v_{t,i} = v(t, x_{t,i})$ is the velocity of the atom $x_{t,i}$. 

Another formulation inspired by the Modica–Mortola (MM) framework was proposed by \citep{oudet2011modica}. It replaces the singular, non-smooth branched transport problem with a sequence of regularized, elliptic energy functionals $M_\lambda^{\alpha}$ defined over the more regular space of $\mathcal{H}^1$ vector fields. For a given $F(x)$, the approximating functional is as follows:
\begin{equation}
    M_\lambda^{\alpha}(F) = \lambda^{\gamma_1} \int_{\Omega} |F(x)|^{\alpha} dx + \lambda^{\gamma_2} \int_{\Omega} |\nabla F(x)|^2 dx,
    \label{eq:modica_mortola_approx}
\end{equation}
where  $\lambda > 0$ is a small parameter. The exponents $\alpha$, $\gamma_1$, and $\gamma_2$ are derived directly from the transport dimension $d$ and the cost exponent $\alpha$. But this method is practically \underline{unstable} and difficult to scale for real-world generative modeling applications. Please see Appendix \ref{appendix:MM}.

\begin{tcolorbox}[colback=RoyalBlue!10, colframe=black, boxrule=0.1mm]
\textit{\textbf{Takeaway}: The limitations of current branched methods create a bottleneck for generative modeling. The discrete formulation of~\cite{xia2003optimal} requires intractable optimization over 1-dimensional paths, while the continuous formulation in~\citep{brasco2011benamou} involves singular measures that are ill-suited for sampling. It remains unclear how to perform Monte-Carlo estimation over these densities. Additionally, we found the Modica-Mortola approximation to be numerically unstable in practice and restricted by its definition as a static field. } 
\end{tcolorbox}

Collectively, these failures motivate our proposal: a simpler, robust framework capable of generating branching structures without these optimization pitfalls.



\section{Method}
In this section, we propose an alternative approach that is simpler than the existing methods. In our method, \underline{we are not} solving BOT exactly. We borrow the branching preference and propose a novel velocity-field formulation that is compatible with neural ODEs. We begin by examining the properties of the flux-based branched formulation. Then we show that similar functionals in the velocity space can provide branching structures, too. Let us first decompose the components of the MM approximation \eqref{eq:modica_mortola_approx}:

\textbf{Concave Flux Term} ($\int |F|^\alpha$). This term forces the transport density to concentrate on a lower-dimensional set, promoting sparsity rather than diffusing throughout the space. Minimizing a concave power encourages $F$ to be 0 (empty space) or concentrated on a lower-dimensional manifold, effectively forming the transport network. 

\textbf{Dirichlet Regularizer} ($\int |\nabla F|^2$). This term penalizes sharp transitions, preventing paths from collapsing into infinitely thin Dirac masses. Forces the flux to spread over a strip of finite width $A$, creating a \emph{corridor}. Without this regularization, the concave flux term would prefer infinitely thin, infinitely dense paths, which are ill-defined in the functional space.

\textbf{The Combined Effect.} Together, these terms create a well-posed tradeoff. The concave term tries to compress the mass into singular lines to maximize efficiency, while the Dirichlet term acts as an opposing \emph{pressure} that expands these lines into smooth channels. The resulting equilibrium is a stable, branched network with branches of finite thickness

Motivated by this framework, we propose a velocity-driven objective that shifts the nonlinearity from the flux field to the velocity field.

\subsection{Y-Shaped Velocity Field}

We translate the branching requirements into an optimization problem over dynamic flow trajectories. Unlike static path-based formulations, our objective depends explicitly on the instantaneous velocity field $v_t$ and density $\rho_t$. The loss function is defined as:
\begin{align}
    \mathcal{L}(\rho,v) = \inf_{\rho,v} & \underbrace{\int_0^1 \!\! \int_{\Omega} \rho_t(x) |v_t(x)|^\alpha \, dx dt}_{\mathcal{T}(\rho,v)} \nonumber \\
    & + \lambda \underbrace{\int_0^1 \!\! \int_{\Omega} \rho_t(x) \|\nabla v_t(x)\|^2 \, dx dt}_{\mathcal{C}(\rho,v)}.
    \label{eq:velocity_objective}
\end{align}
This formulation is fully compatible with standard Neural ODE solvers, which naturally satisfy the continuity equation by advecting particles through the learned field $v_t$. Although distinct from the mass-flux approach, this objective drives the emergence of Y-shaped structures through the interplay of two competing dynamic mechanisms: \emph{Cohesion} and \emph{time} efficiency. We analyze this interplay by addressing three fundamental questions regarding the flow behavior.

\subsection*{1. What mechanism enforces the initial collective motion?}
The formation of a shared \emph{trunk} is driven by the \emph{cohesion} term $\mathcal{C}$, which enforces regularity by penalizing spatial changes in the velocity field ($\nabla v$).

\textbf{Proposition 1} (Zero-Cost of Uniform Motion).
\emph{If the velocity field $v(x,t)$ represents a spatially uniform translation (rigid body motion without rotation) on the support of $\rho_t$, the Cohesion Cost density is zero.}

The gradient $\nabla v$ measures the difference in velocity between neighboring points. When the entire probability mass moves with the exact same velocity vector, this gradient vanishes. Consequently, the objective function assigns zero penalty to this state, creating a strong preference for the mass to remain aggregated and translate as a single unit for as long as possible.

\subsection*{2. What acts as the barrier to premature separation?}
To satisfy the boundary conditions, the mass must eventually separate to reach different targets (e.g., $y_1$ and $y_2$). However, separating a continuous density requires the velocity field to diverge in opposing directions.

\textbf{Proposition 2} (The Cost of Separation).
\emph{If a continuous velocity field separates the mass into distinct directions over a transition region of width $\epsilon$, the Cohesion Cost scales as $\mathcal{O}(1/\epsilon)$.}

To split the flow, velocity vectors at the same time step must point in different directions. If this change in direction occurs over a very short distance $\epsilon$, it creates a steep slope, resulting in a large gradient magnitude. Since the cost function effectively integrates the squared slope ($\|\nabla v\|^2$), narrowing the gap $\epsilon$ causes the energy cost to explode. This acts as a barrier that prevents the trajectories from splitting until they are spatially distant enough to do so smoothly.

\subsection*{3. How does the transport cost enable delayed branching?}
If the system delays splitting to avoid the cohesion penalty, the particles must traverse the remaining distance in a reduced time interval. This requires traveling at higher speeds.

\textbf{Proposition 3} (Time-Compression).
\emph{For $\alpha \in (0,1)$, the cost of traveling a fixed distance decreases as the duration of the travel decreases. Specifically, the functional favors high-velocity, short-duration transport over constant-velocity transport.}

The transport cost $|v|^\alpha$ increases sub-linearly with speed when $\alpha < 1$. This implies an economy of scale with respect to time: doubling the speed increases the instantaneous cost by a factor less than two, while halving the travel time. Therefore, the total integral over time is lower for fast, short bursts than for slow, continuous motion. We term this \emph{time-compression}, as it allows the system to offset the cost of waiting in the trunk.

\textbf{Summary.} The system minimizes total energy by balancing two regimes: the zero-cost uniform motion favored by $\mathcal{C}$ (Prop. 1) and the high cost of splitting (Prop. 2). Branching occurs only when the energetic gain from time-compression (Prop. 3) sufficiently dominates the splitting penalty. This trade-off defines a critical branching time $\tau^* > 0$.

\textbf{Lemma 2} (Existence of Optimal Branching Time).
\emph{For a sufficient cohesion weight $\lambda$ and $\alpha \in (0,1)$, the optimal branching time satisfies $\tau^* > 0$, implying that a Y-shaped trajectory is energetically superior to a V-shaped trajectory ($\tau=0$).}
See Appendix \ref{sec:intuition_proofs} for full proofs.

\section{Practical Implementation}

We now describe a simple and scalable training scheme based on neural ODEs and a Monte Carlo (MC) approximation of the velocity-power action, now augmented with the cohesion regularization term. Throughout this section, we fix the time horizon to $[0, 1]$.

We parameterize the velocity field by a neural network $v_{\theta}: \Omega \times [0, 1] \rightarrow \mathbb{R}^d$ and evolve samples via the ODE \ref{sec:premilinaries}: $\frac{dx_t}{dt} = v_{\theta}(x_t, t), \quad x_{t=0} \sim \mu_0.$ Let $\Phi_t$ denote the flow map. The time-$t$ distribution of particles is the push-forward $\rho_t = (\Phi_t)_{\#} \mu_0$; i.e., if $x_0 \sim \mu_0$ then $x_t = \Phi_t(x_0)$ is a draw from $\rho_t$. This simple fact justifies using equal-weight particles to approximate expectations under $\rho_t$; no importance weights are introduced or needed.

\textbf{Objective.} The total action we minimize combines the sub-linear transport cost with the cohesion regularizer, as motivated in Section 4. We minimize:
\begin{equation}
    Y^{\alpha}(\theta) = \int_{0}^{1} \mathbb{E}_{x \sim \rho_t} \left[ |v_{\theta}(x, t)|_2^{\alpha} + \lambda | \nabla_x v_{\theta}(x, t) |^2 \right] dt,
    \label{eq:objective}
\end{equation}
where $\alpha \in (0, 1)$ is the branching exponent and $\lambda > 0$ is the cohesion weight that matches the formulation of the method. The term $\| \nabla_x v_{\theta} \|_F^2$ represents the Dirichlet energy of the velocity field, which is efficiently computed using automatic differentiation.

\textbf{Time discretizations and MC estimator.} Let $0 = t_0 < t_1 < \dots < t_K = 1$ be a time grid with steps $\Delta t_k = t_k - t_{k-1}$ (in the uniform case, $\Delta t_k \equiv \Delta t = 1/K$). For a mini-batch $\{x_0^{(i)}\}_{i=1}^N \sim \mu_0$, we integrate our neural ODE along the grid to obtain states $x_{t_k}^{(i)} = \Phi_{t_k}(x_0^{(i)})$. The consistent estimator of Equation \eqref{eq:objective} is then:
\begin{equation}
\begin{aligned}
    \hat{Y}^{\alpha}(\theta) &= \frac{1}{N} \sum_{i=1}^{N} \sum_{k=1}^{K} \Delta t_k \, |v_{\theta}(x_{t_{k-1}}^{(i)}, t_{k-1})|_2^{\alpha} \\
    &\quad + \lambda \frac{1}{N} \sum_{i=1}^{N} \sum_{k=1}^{K} \Delta t_k \, | \nabla_x v_{\theta}(x_{t_{k-1}}^{(i)}, t_{k-1}) |^2.
\end{aligned}
\label{eq:estimator}
\end{equation}
In the case of the common uniform grid, this simplifies the outer time summation to $\frac{1}{K} \sum_{k}$. If an adaptive ODE solver is used, we evaluate its internal time points and accumulate the corresponding $\Delta t_k$. Under exact integration, independence of $\{x_0^{(i)}\}$ and mesh refinement ($N, K \rightarrow \infty$), $\hat{Y}^{\alpha}(\theta) \rightarrow Y^{\alpha}(\theta)$ by the law of large numbers and Riemann sums.

\textbf{Boundary Matching.} To enforce $\rho_{t=1} = \mu_1$, we add a differentiable boundary loss that compares the empirical law of the final particles with $\mu_1$. Given mini-batches $X_1 = \{x_{t=1}^{(i)}\}_{i=1}^{N_X}$ (push-forward samples) and $Y = \{y^{(j)}\}_{j=1}^{N_Y} \sim \mu_1$, we use the entropic Sinkhorn divergence (bias-corrected) \cite{cuturi2013sinkhorn, feydy2019interpolating}, with cost $c(x,y) = \|x-y\|_2^2$. Where $OT_{\epsilon}$ is the entropic OT cost computed by Sinkhorn iterations (regularization $\epsilon > 0$).

\textbf{Final Loss.} Our total training objective is:
\begin{equation}
    \mathcal{L}(\theta) = \hat{Y}^{\alpha}(\theta) + \lambda_{sink} \mathcal{L}_{sinkhorn}(\theta),
\end{equation}
where $\lambda_{sink} > 0$ balances the boundary matching constraint against the trajectory action. We differentiate through the ODE solver (using standard backpropagation or the adjoint method) and optimize $\theta$ with stochastic gradient methods using fresh mini-batches from $\mu_0$ and $\mu_1$. In practice, for efficient Jacobian computation, the Hutchinson trick was applied \cite{grathwohl2018ffjord}. 

\section{Related Work}
\label{sec:related}
Branching transport problems were studied in discrete settings. 
Classical approaches include the Euclidean Steiner Tree (ESTP), where branch points are geometrically optimized under flow-independent costs, and early geometric constructions for single-source 
branched optimal transport (BOT) \citep{buttazzo2003optimal,bernot2005traffic, maddalena2003variational}. More recently, \citep{lippmann2022theory} introduced an approximate solver that decouples topology search from geometry optimization: given a tree topology, the positions of the branches are optimized through leaf-elimination in $O(nd)$ time, while a rewiring procedure explores the topology. Their method works only for \textit{discrete} settings. 

Beyond discrete graph solvers, continuous stochastic frameworks have also been proposed. \citep{tang2025branched} extend Schrödinger Bridge Matching (BSBM) to the \emph{branched} case by learning divergent drift and growth fields that transport mass from a common source to multiple targets. This method produces continuous branched trajectories, but is not based on branched transport theory.

\section{Experiments}
\subsection{Branching Illustrations}
\label{sec:experiments}
\begin{figure*}[t!]
  \centering
  \setlength{\tabcolsep}{4pt} 

  \begin{tabular}{c c c c c}
    & 2 branches & 4 branches & 6 branches & 18 branches \\[0.5em]

    \rotatebox{90}{ Flow Matching} &
    \includegraphics[width=.21\textwidth]{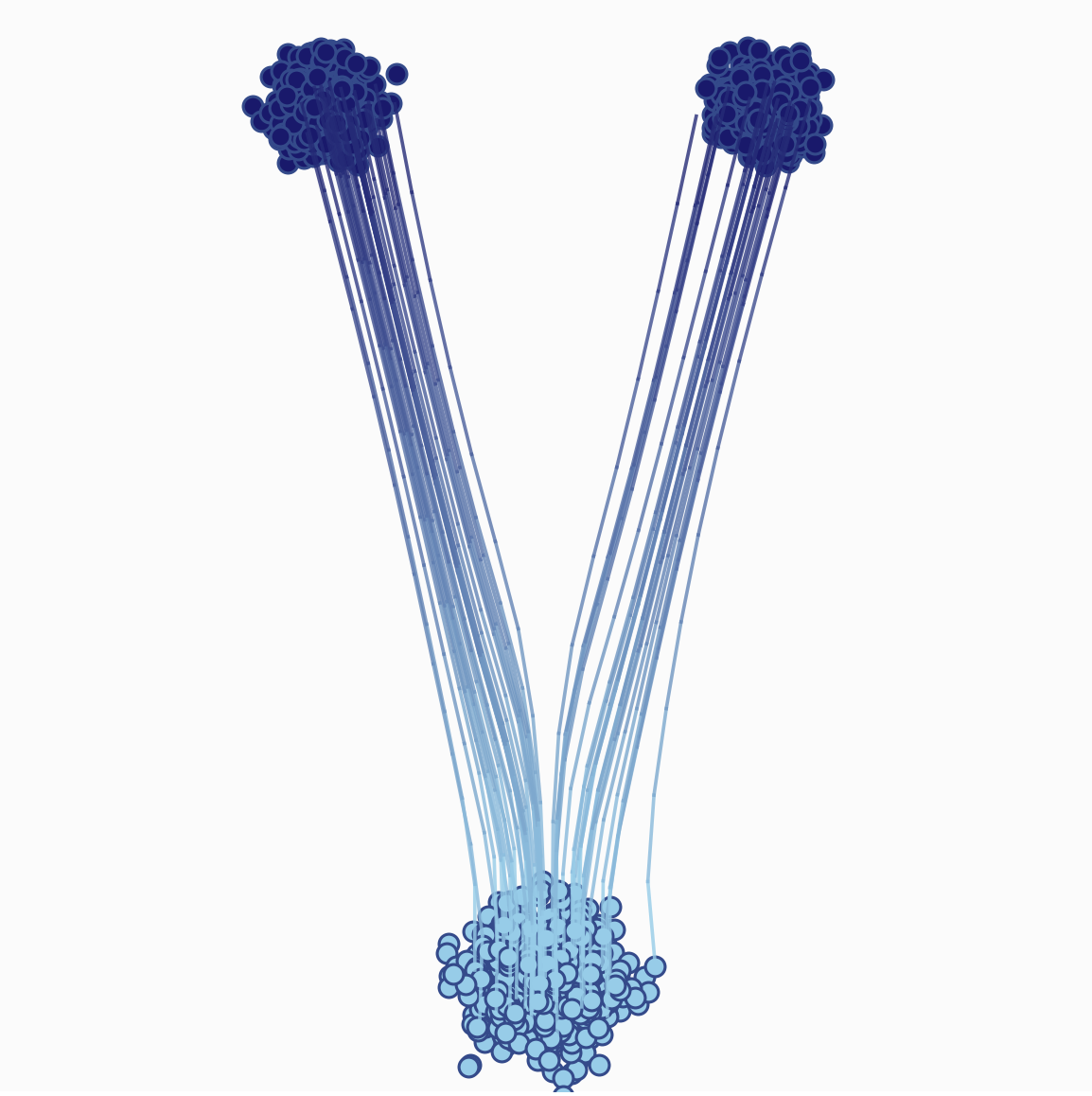} &
    \includegraphics[width=.21\textwidth]{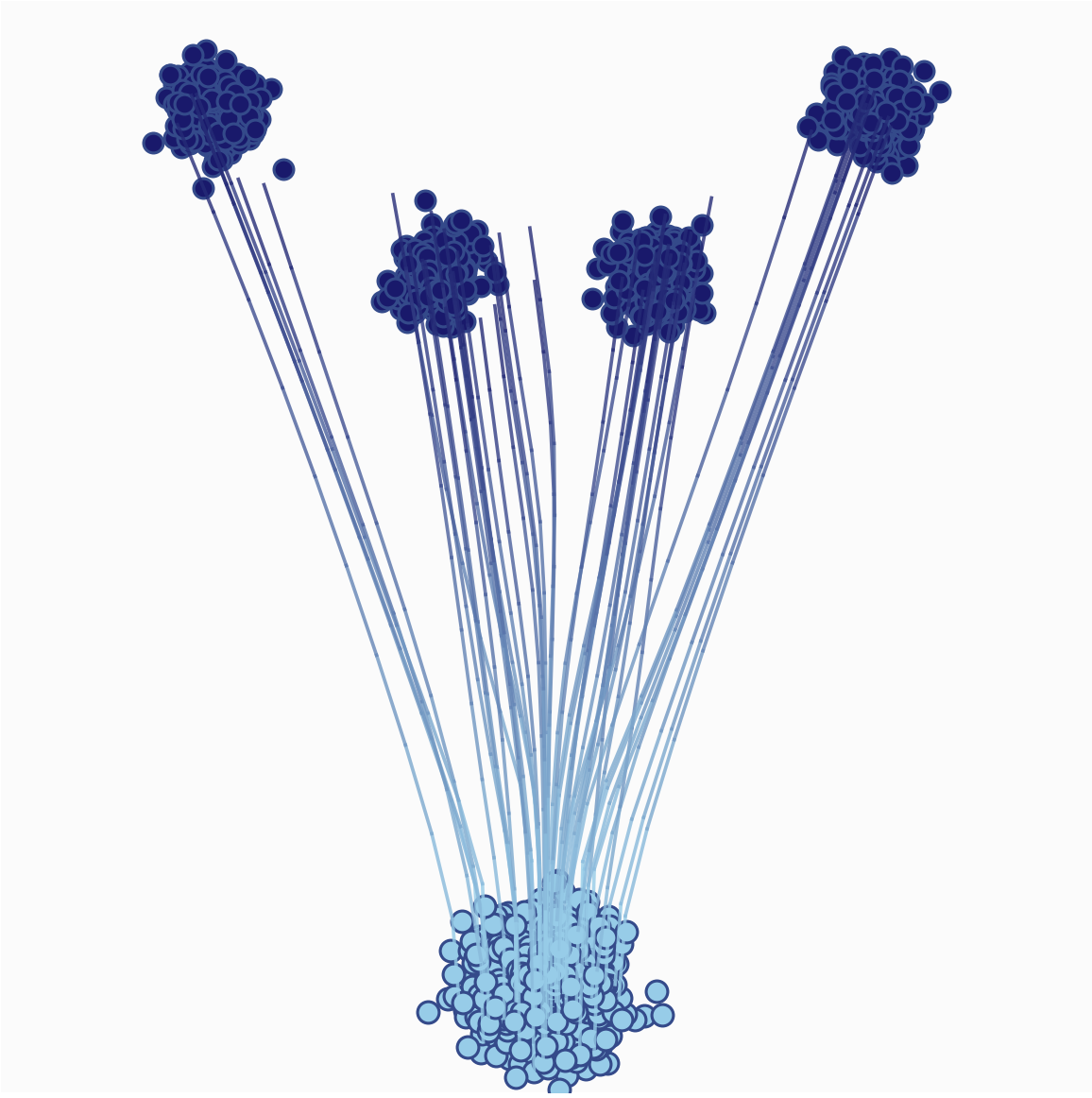} &
    \includegraphics[width=.21\textwidth]{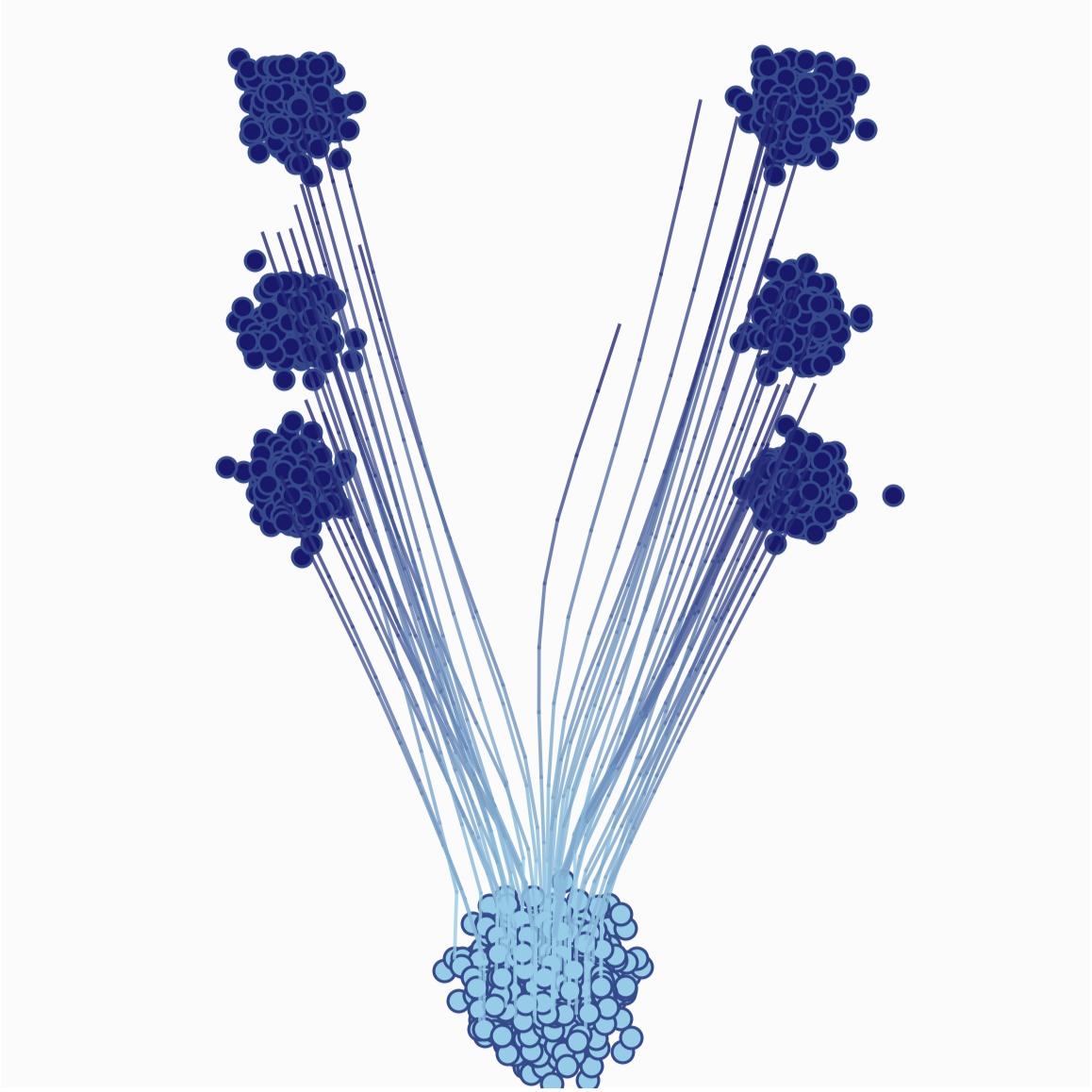} &
    \includegraphics[width=.21\textwidth]{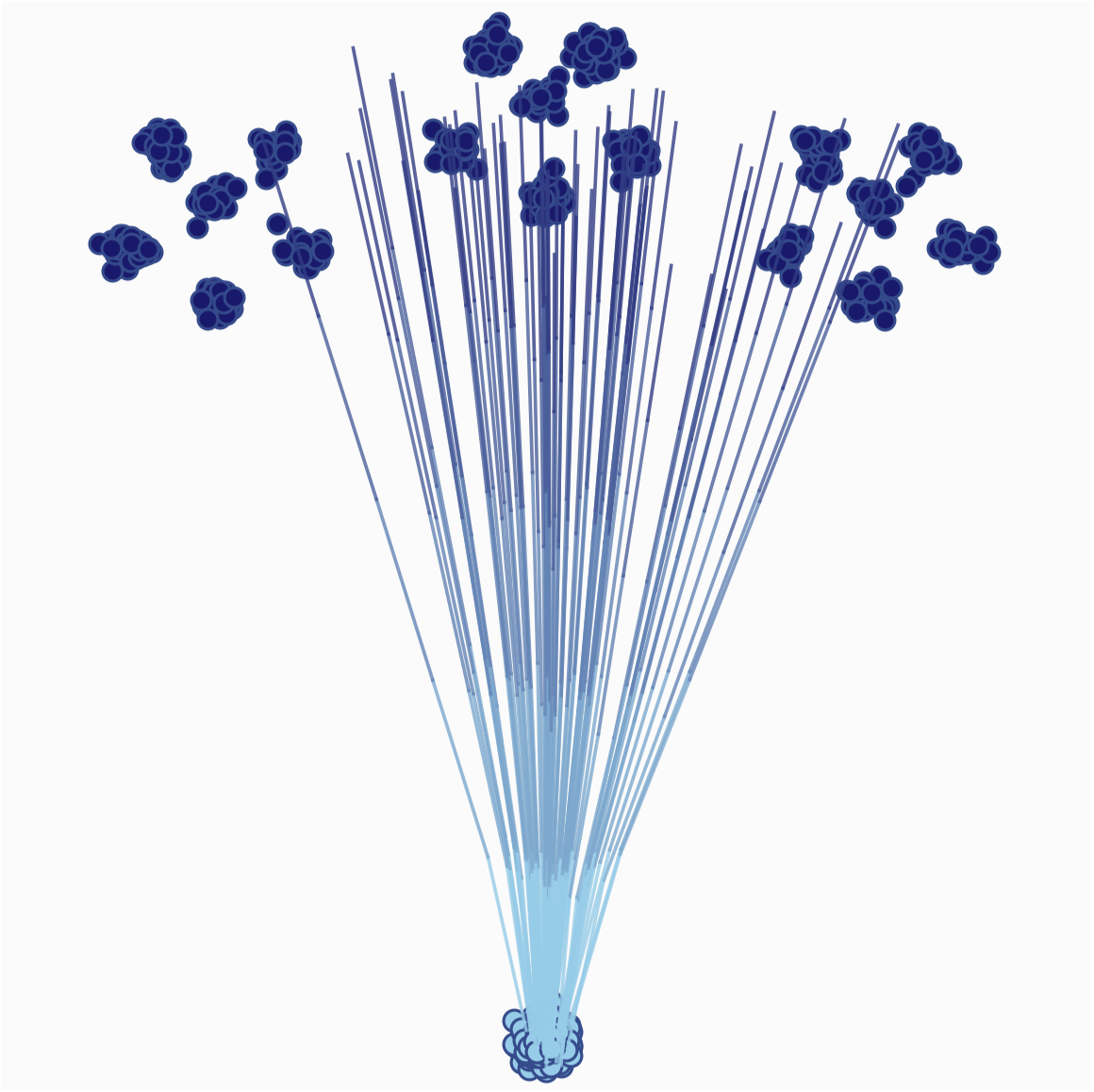} \\[1em]

    \rotatebox{90}{\qquad Y-Flows} &
    \includegraphics[width=.21\textwidth]{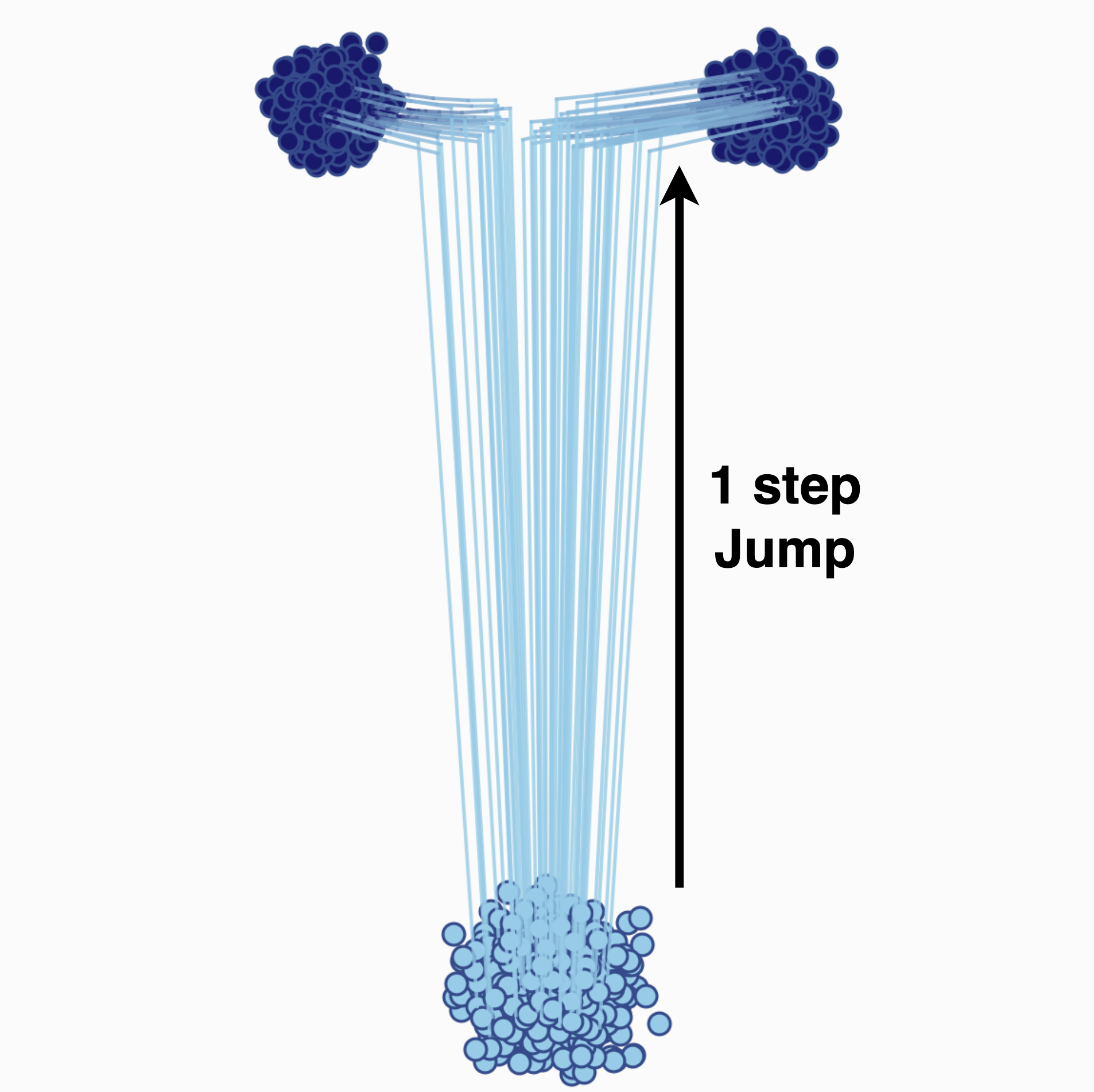} &
    \includegraphics[width=.21\textwidth]{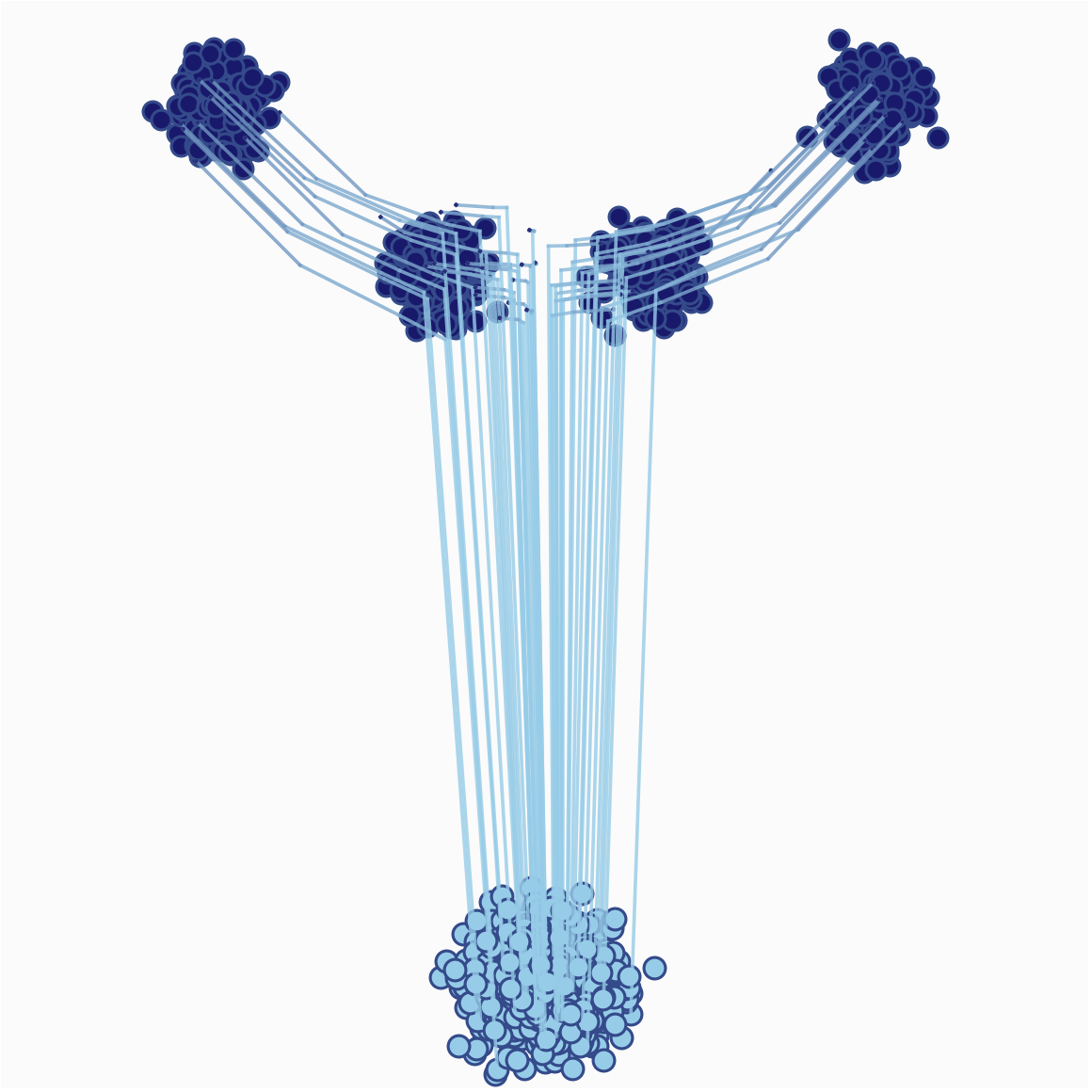} &
    \includegraphics[width=.21\textwidth]{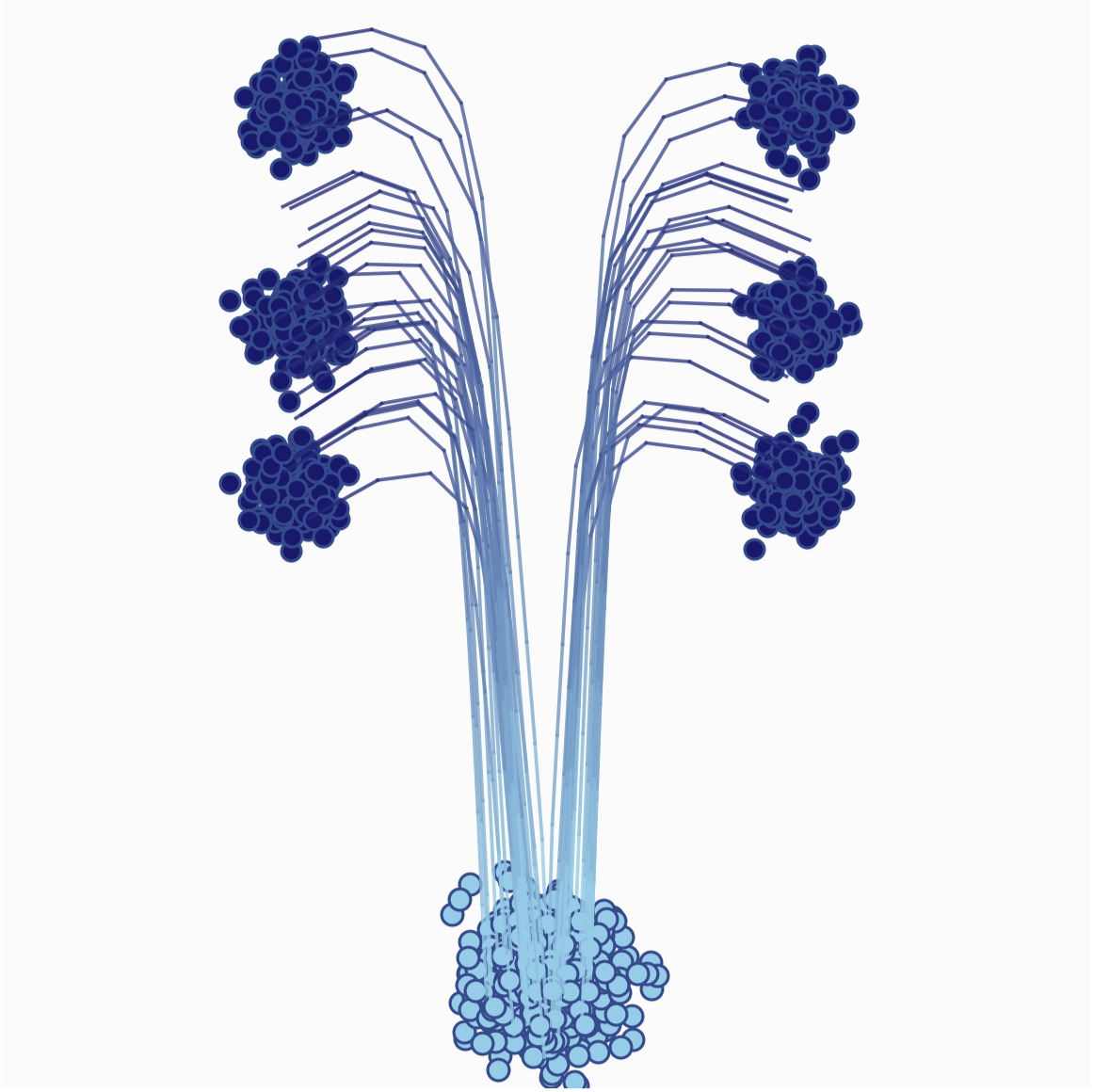} &
    \includegraphics[width=.21\textwidth]{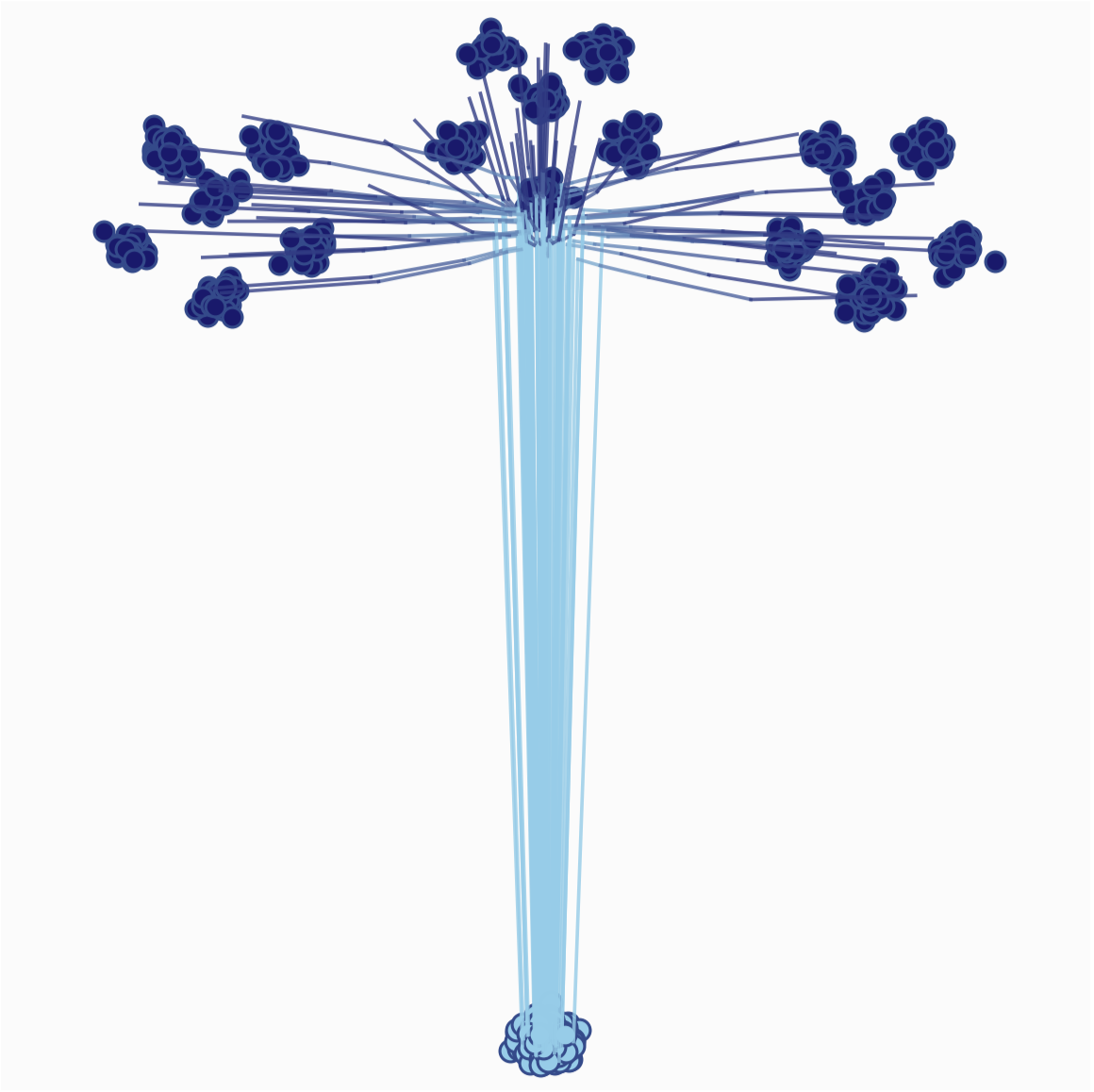} \\
  \end{tabular}

  \caption{Comparison of Gaussian mixture toy tasks. Each \textbf{column} is a task 
  (T-shape, 4,6,18 Branches). Each \textbf{row} is a method: 
  top = Flow Matching (FM), bottom = Y-Flows (ours). The color gradient represents the flow steps. A monotone color indicates that the number of steps on this region was equal to 1. As shown in the 2, 4, and 18-branch cases, our model initially made a significant jump toward the target before splitting the mass. Subsequent movement involved reaching the targets via small almost zero size steps.}
  \label{fig:gaussian-mixtures-horizontal}
  \vspace{-3mm}
\end{figure*}
\textbf{Gaussian Mixtures}. We compare how two flow-based methods transport mass from a single source distribution to a highly multi-modal target. The target is a mixture of $K$ groups (“branches”) arranged at the top; the source is a single group at the bottom. We train the same-size models with (i) standard FM and (ii) Y-Flows. Each curve shows the trajectory of a sample from source (light) to destination cluster (dark).
Both models use the same time-conditioned MLP $v_\theta(x, t)$: 64-$d$ time embedding $\rightarrow$ 3×256 SiLU layers $\rightarrow$ 2-D output; integrated with 10 fixed Euler steps over $t\in [0,1]$. Training is identical: Adam ($lr=10^{-3}$, batch 256, 10k iterations, seed 42. FM uses the standard flow-matching objective along linear source–target couplings. Y-Flows maintains the same backbone, but increases the loss with a branched-transport prior: endpoint OT/Sinkhorn term weight $\lambda_{sink} \approx 5$. For the results, see Figure \ref{fig:gaussian-mixtures-horizontal}. As shown, Y-Flows discovers a shared \emph{trunk} that later splits into branches, producing short, structured, tree-like transport.

To take advantage of the fast convergence properties, we propose an early stopping mechanism for the neural ODE solver. Specifically, we terminate the integration if the displacement norm between consecutive steps, $\|x_{t+1} - x_t\|_2$, falls below a threshold of $\delta = 1 \times 10^{-3}$. This criterion effectively detects when the particles have settled in their target modes, allowing the model to dynamically adapt its computational budget. Empirically, we observe that on the Gaussian mixtures, Y-Flows converged in an average of 3 steps for the 2-branch case, 5 steps for 4 branches, 10 steps for 6 branches, and 4.5 steps for the 18-branch setting. In contrast, the Flow Matching baseline consistently utilized a fixed budget of 10 integration steps to achieve comparable transport, highlighting our method's ability to concentrate mass movement into fewer updates.
\begin{figure}[h!] %
    \centering
\includegraphics[width=0.5\textwidth]{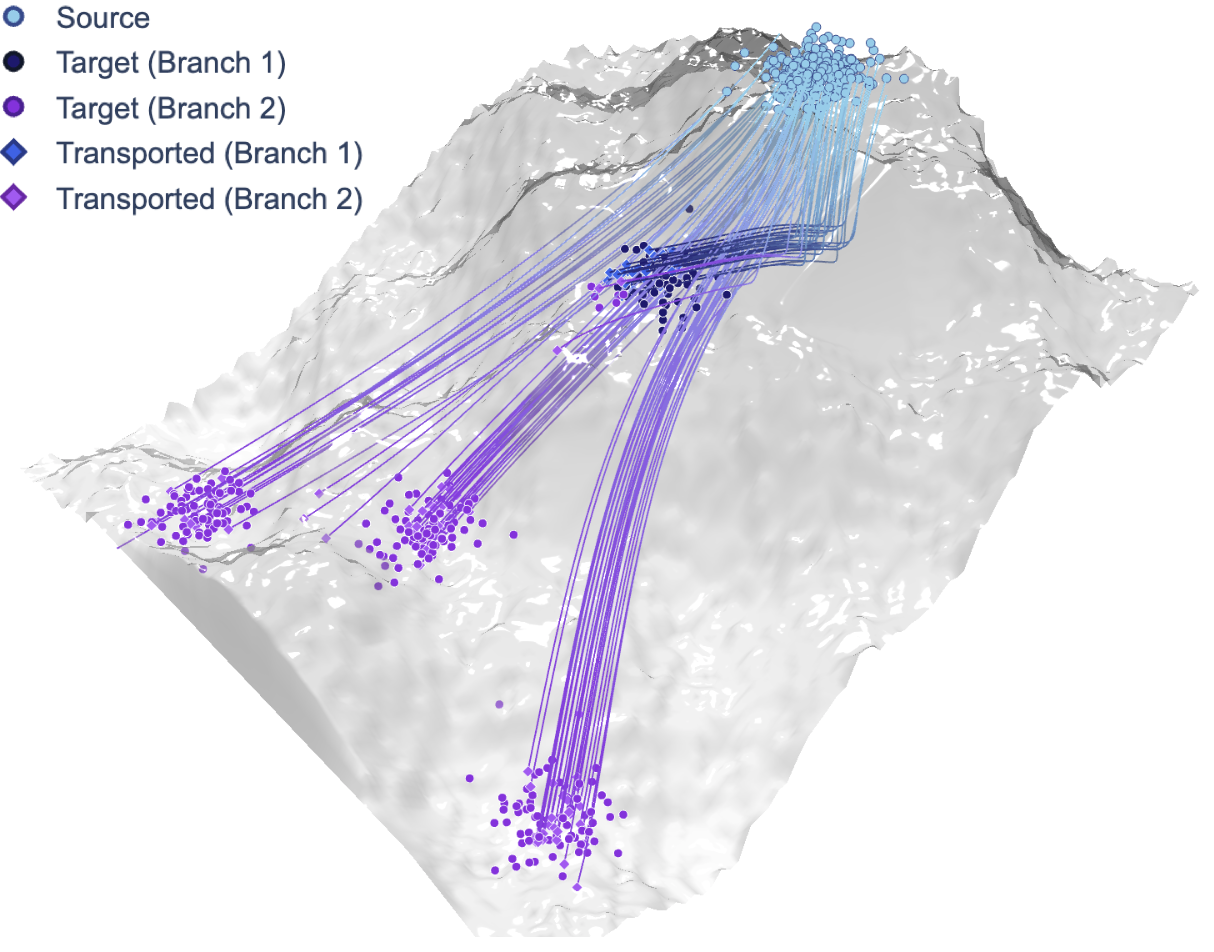}
    \caption{\emph{Result of Y-Flows on LiDAR dataset}.}
    \vspace{-6mm}
    \label{fig:Y-shaped_Lidar}
\end{figure}

\textbf{LiDAR Surface Navigation}. The purpose of this experiment is to test whether a branched OT model \textit{can learn when and where to split while staying confined to a real 3D surface}. For this experiment, we include a potential energy penalty on our loss as in \citep{liu2023generalized,tang2025branched}. We transport a single source to four target clouds laid out on an airborne LiDAR terrain, which adds curvature, uneven sampling, and natural surface corridors that synthetic mixtures do not capture. 

Each sample is a 3D point $(x,y,z)$ with optional attributes (intensity, return number, total returns, scan angle, class). After filtering non-ground classes, removing outliers with a radius/k-NN check, and Poisson-disk (or voxel-grid) subsampling, we retain ~5k points that preserve ridges, slopes, and basins. Coordinates are centered and scaled to a unit box (optionally PCA-aligned). We estimate per-point normals via local PCA and build a k-NN surface graph $(k \in [8,16])$ with Euclidean edge weights as a geodesic proxy. During optimization, states are projected back to the nearest tangent patch to stay on-manifold. A single source distribution sits on a lower slope; four target clouds are disjoint regions on ridges/basins (indices provided for reproducibility). 

The method infers a three-junction topology, places the splits at terrain transitions (ridge breaks/valley entries), and keeps trajectories on-surface, reconstructing all four targets with tight endpoint fits. This shows that the approach handles surface-constrained, branched transport and is applicable in real-world tasks; see Figure~\ref{fig:Y-shaped_Lidar} for details.

\subsection{Biology Data}
\begin{figure*}[t!]
    \centering
    \begin{subfigure}{0.48\textwidth}
        \centering
        \includegraphics[width=\linewidth]{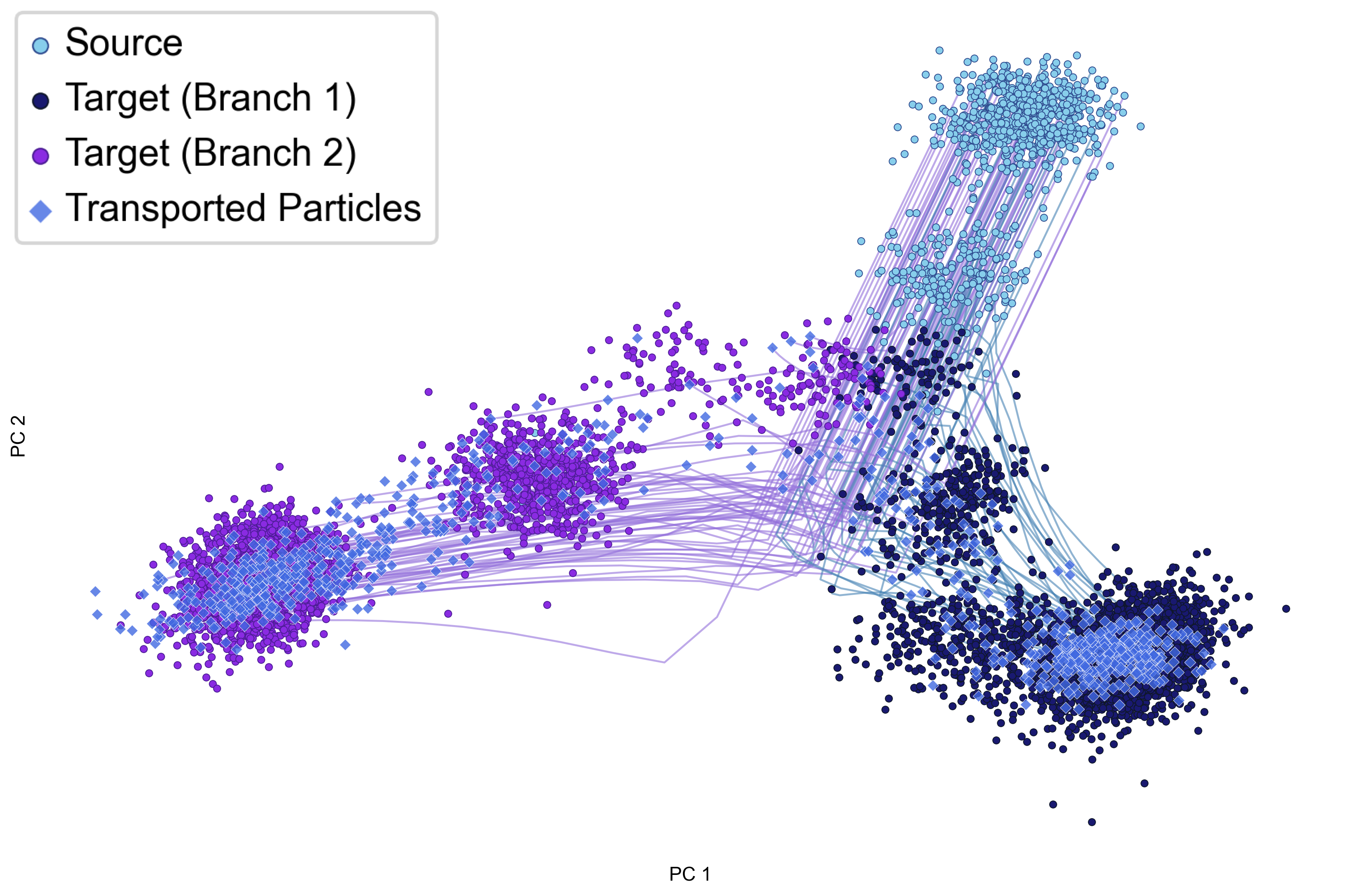}
        \caption{$\alpha=0.5$.}
    \end{subfigure}
    \begin{subfigure}{0.48\textwidth}
        \centering
        \includegraphics[width=\linewidth]{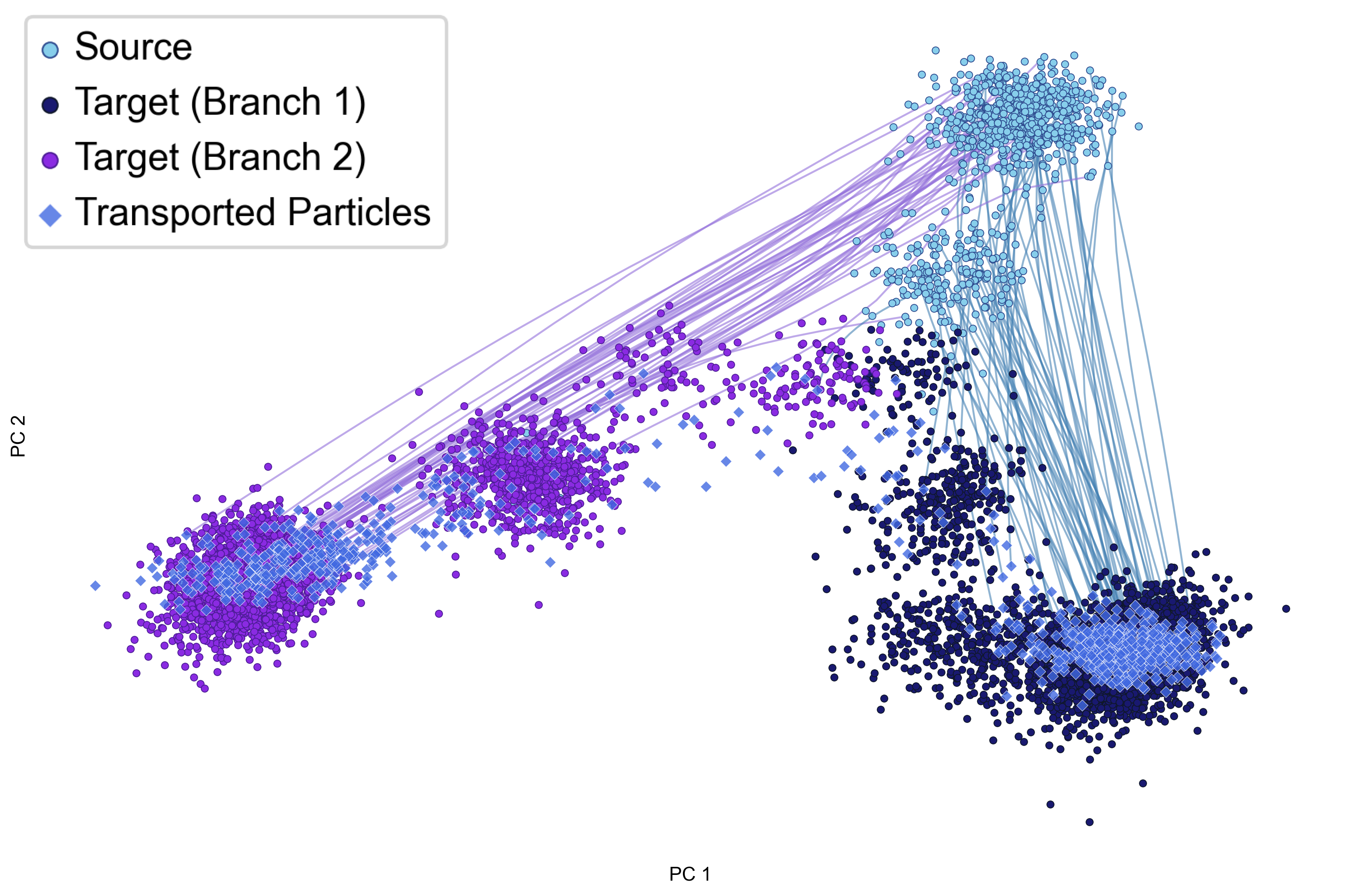}
        \caption{$\alpha=1.0$.}
    \end{subfigure}
    \caption{\emph{PCA projected results of Y-Flows on Tedsim dataset (50D).}}
    \vspace{-3mm}
    \label{fig:Y-shaped_tedsim}
\end{figure*}
\textbf{Cellular Differentiation}. To further validate our method, we use the Tedsim dataset \cite{pan2022tedsim} as a controlled reference point with known ground truth dynamics. Tedsim simulates a cellular differentiation process by modeling cell division from a root cell, generating both gene expression profiles and heritable lineage barcodes. This provides a complete record of a simple, branching differentiation process.

We apply our method to a specific Tedsim scenario, modeling the transport from a single progenitor cell to two distinct terminal states. This controlled experiment allows us to quantitatively assess our method's ability to accurately reconstruct branching trajectories where the true pathways are known beforehand. See Figure \ref{fig:Y-shaped_tedsim} and Table \ref{tab:comparison_joint}. The experiments highlight our model's effectiveness in reconstructing branching trajectories. We observed that sub-linearity exponents of $\alpha =0.5$  consistently produced well-defined bifurcating structures. We compared different approaches using the following metrics:  W1/W2: Wasserstein distance with ground metrics, see Section \ref{subsec:branched_ot_bb}. RBF-MMD: Maximum Mean Discrepancy using a Radial Basis Function kernel with a median heuristic bandwidth\cite{mmd}. See Table \ref{tab:comparison_Tedsim_extended} for results in higher dimensions.

\textbf{Single-cell RNA}. We now evaluate the performance of our method on the 50D Paul15 dataset. This dataset \citep{paul2015transcriptional} contains single-cell RNA sequencing data from approximately 2,730 myeloid progenitor cells, with expression profiles for roughly 1,000 highly variable genes. 

The dataset captures various cell types in the hematopoietic system, including early progenitor cells, monocytes, and neutrophils. This dataset represents cellular states during myeloid differentiation and serves as a benchmark for studying developmental trajectories and cell fate decisions in hematopoiesis. We model transport from early progenitor cells to mature, differentiated cell types (monocytes and neutrophils). 
For comparison with other methods such as Branched SBM \citep{tang2025branched} and FM \citep{lipman2022flow}, see Table \ref{tab:comparison_Tedsim_extended} and  \ref{tab:comparison_paul_extended}. 
\begin{table}[t!]
    \centering
    \caption{Comparison across methods on Tedsim dataset across embedding dimensions.}
    \label{tab:comparison_Tedsim_extended}
    \scriptsize
    \begin{tabular}{llccccc}
        \toprule
         & Metric & BSBM & CNF & CFM & FM & Y-Flows \\
        \midrule
        \multirow{3}{*}{50D} 
            & $W_1$   & 17.72 & 13.76 & 12.33 & 12.09 & \textbf{10.48} \\
            & $W_2$   & 17.96 & 13.81 & 12.44 & 12.17 & \textbf{10.72} \\
            & RBF-MMD & 0.63 & 0.51 & 0.11 & 0.15 & \textbf{0.10} \\
        \midrule
        \multirow{3}{*}{150D} 
            & $W_1$   & 20.34 & 18.84 & 17.45 & 18.60 & \textbf{14.29} \\
            & $W_2$   & 20.51 & 18.88 & 17.54 & 19.79 & \textbf{14.42} \\
            & RBF-MMD & 0.48 & 0.29 & 0.13 & 0.14 & \textbf{0.13} \\
        \midrule
        \multirow{3}{*}{250D} 
            & $W_1$   & 23.47 & 22.48 & 20.07 & 22.80 & \textbf{17.37} \\
            & $W_2$   & 23.71 & 22.51 & 20.14 & 22.94 & \textbf{17.50} \\
            & RBF-MMD & 0.55 & 0.18 & 0.12 & 0.16 & \textbf{0.11} \\
        \midrule
        \multirow{3}{*}{500D} 
            & $W_1$   & 29.46 & 28.88 & 22.60 & 24.97 & \textbf{20.99} \\
            & $W_2$   & 29.63 & 28.91 & 22.68 & 25.04 & \textbf{21.05} \\
            & RBF-MMD & 0.51 & 0.15 & 0.12 & 0.13 & \textbf{0.12} \\
        \bottomrule
    \end{tabular}
\end{table}
\begin{table}[t!]
    \centering
    \caption{Comparison across methods on Single-Cell RNA dataset across embedding dimensions.}
    \label{tab:comparison_paul_extended}
    \scriptsize
    \begin{tabular}{llccccc}
        \toprule
         & Metric & BSBM & CNF & CFM & FM & Y-Flows \\
        \midrule
        \multirow{3}{*}{50D} 
            & $W_1$   & 17.65 & 8.69 & 8.22 & 8.16 & \textbf{7.06} \\
            & $W_2$   & 18.11 & 9.14 & 8.52 & 8.41 & \textbf{7.28} \\
            & RBF-MMD & 0.47  & 0.17 & 0.13 & 0.14 & \textbf{0.12} \\
        \midrule
        \multirow{3}{*}{150D} 
            & $W_1$   & 25.68 & 16.73 & 13.90 & 13.81 & \textbf{12.90} \\
            & $W_2$   & 26.13 & 17.13 & 14.23 & 14.17 & \textbf{12.68} \\
            & RBF-MMD & 0.35  & 0.15 & 0.10 & 0.12 & \textbf{0.09} \\
        \midrule
        \multirow{3}{*}{250D} 
            & $W_1$   & 30.36 & 22.13 & 17.67 & 18.07 & \textbf{16.27} \\
            & $W_2$   & 30.77 & 22.49 & 17.95 & 18.36 & \textbf{16.46} \\
            & RBF-MMD & 0.31  & 0.15 & 0.11 & 0.12 & \textbf{0.11} \\
        \midrule
        \multirow{3}{*}{685D} 
            & $W_1$   & 33.81 & 33.45 & 25.26 & 25.68 & \textbf{23.28} \\
            & $W_2$   & 34.07 & 33.56 & 25.45 & 25.88 & \textbf{23.44} \\
            & RBF-MMD & 0.39  & 0.20 & 0.14 & 0.16 & \textbf{0.11} \\
        \bottomrule
    \end{tabular}
    \vskip-8pt
\end{table}

\section{Image data}
\textbf{FFHQ dataset.} To test our model in higher dimensions we run experiments in 512 D latent space of a pretrained ALAE model on FFHQ 1024x1024 dataset. The ALAE is used \emph{only} for decoding latents to images during evaluation and remains frozen at all times. We load latent vectors paired with gender labels and split them into 60k training and 10k test samples. Let $\mathcal{D}$ denote the latent dimensionality inferred from the data. 

\begin{figure*}
    \centering
    \includegraphics[width=0.99\linewidth]{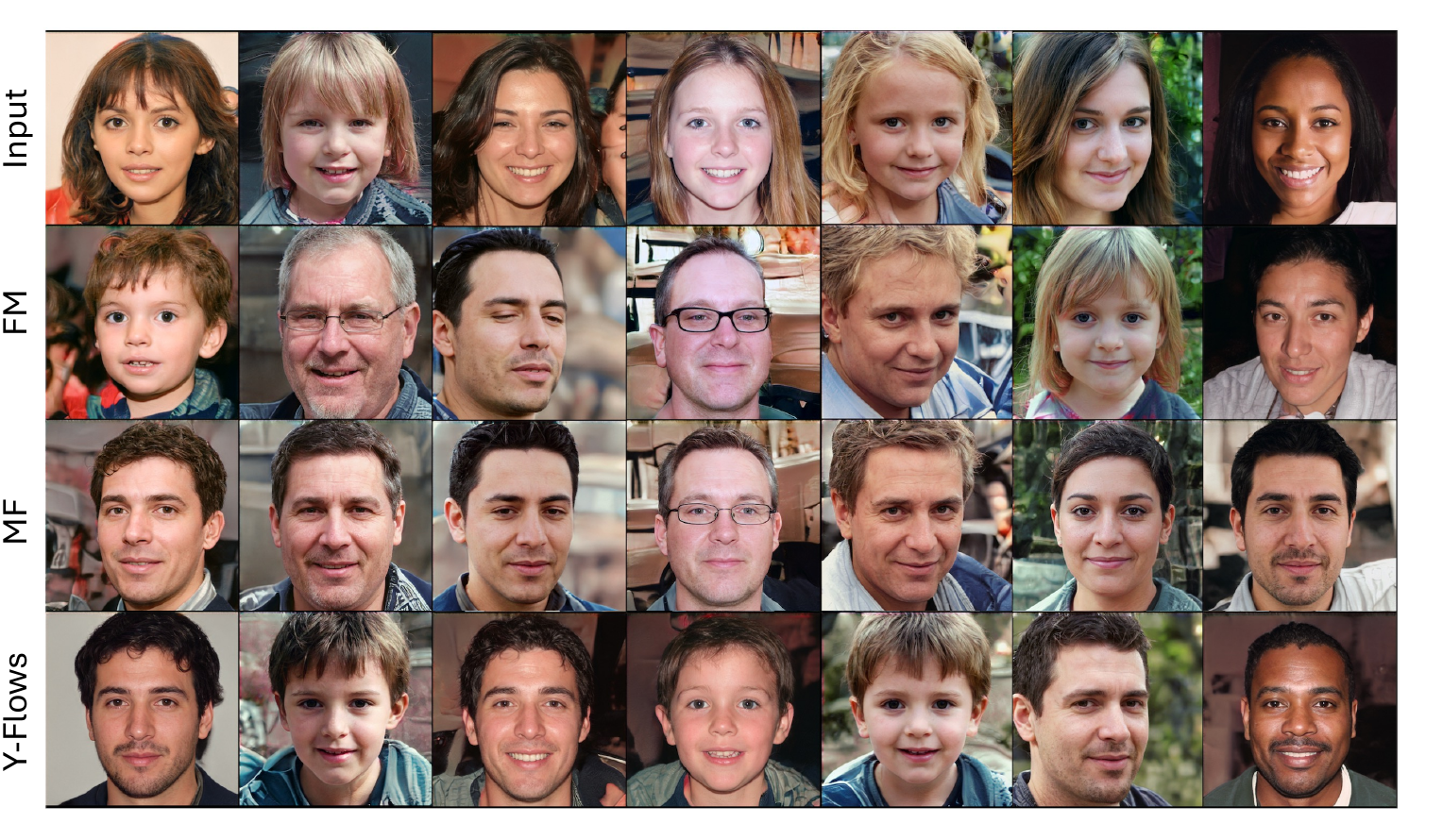}
    \caption{\textit{Female-to-male domain translation in the ALAE latent space. Two-step ODE results. Our method produces compatible results in a higher-dimensional latent space, as do other generative models.}}
    \vspace{-4mm}
    \label{fig:female-male}
\end{figure*}

\textbf{Settings}: \textit{Unconditional generation}: learn a vector field that transports samples from a standard Gaussian prior $\mu_0 \sim \mathcal{N}(0, I_{\mathcal{D}})$ to the empirical distribution of latent data $\mu_1$. \textbf{ 2)} \textit{Domain translation}: learn class-conditional transport from source-class latents to target-class latents (e.g. female $\to$ male); mini-batches for $\mu_0$ and $\mu_1$ are sampled from the respective class-specific training subsets.

\textbf{Architecture}: $v_\theta$ is an MLP with Tanh activations and hidden width 1024. A scalar time embedding (linear layer) is concatenated to $x$ before the MLP.  We use the same architecture for fair comparison for all types of model, except for Mean Flows which requires additional input $r$ to encode the start of sampling trajectory. 

\textbf{Baselines}: We compare with Flow Matching (FM) and Mean Flows (MF) \citep{geng2025mean}. All methods are trained for 100k iterations with a learning rate of $10^{-4}$ using the Adam optimizer.  
\begin{table}[t!]
  \centering
  \vskip -5pt
    \caption{Fréchet distance \textbf{val} vs.\ steps by model ($\downarrow$) on FFHQ.}
   \vskip -5pt
  \label{tab:fd_val_steps_transposed}
  \renewcommand{\arraystretch}{1.12}
  \begin{tabular}{
      S[table-format=2.0]
      S[table-format=2.2]
      S[table-format=2.2]
      S[table-format=2.2]
  }
    \toprule
    {Steps} & {Flow Matching } & {Mean Flows}  & {Y-Flows}\\
    \midrule
    1  & 35.17 & 35.87   & 35.27 \\
    2  & 26.33 & 23.56   & \textbf{23.45} \\
    \bottomrule
  \end{tabular}
\end{table}

\textbf{Results:} Figure~\ref{fig:female-male} visualizes the domain translation results. Top row: samples from the source distribution $\mu_0$ (female faces). Subsequent rows show the corresponding translations in the target distribution $\mu_1$ (male faces) using FM, MF, and Y-Flows. The columns share the same source sample for cross-method comparison. Qualitatively, Y-Flows better preserves key attributes such as apparent age and skin tone. Quantitatively, we report the Fréchet Distance (\textbf{FD}) on FFHQ latents. As shown in Table~\ref{tab:fd_val_steps_transposed}, Y-Flows achieves the best scores and achieves high-quality translations in as few as two steps. See Figure~\ref{fig:generation} for the unconditional generation. This justifies our theoretical assumption of compression of time in our method. 

\textbf{Run-time:} All experiments were conducted on a single NVIDIA RTX A6000 GPU. The average training time per experiment takes $\sim$ 1 hour. To \textit{reproduce} our experiments, refer to the supplementary materials. The code will be open-source. Details on the hyper-parameters used are presented above.

\section{Conclusion}
In this work, we introduced Y-shaped generative flows, a continuous-time framework that addresses the structural limitations of standard V-shaped transport by rewarding shared movement before branching. By minimizing a novel velocity-power action with a sublinear exponent $\alpha \in (0,1)$, we established a theoretical link between scalable neural ODEs and branched structures. We demonstrated that this concave velocity dependence induces a time-compression effect, favoring fast, shared transport along trunks followed by efficient branching. Empirically, our method recovers interpretable lineage structures in biological data, respects geometric constraints in LiDAR navigation, and achieves high-quality image generation with significantly fewer integration steps. Ultimately, Y-Flows offer a theoretically grounded and practically effective approach to learning hierarchy-aware generative trajectories that adapt computational effort to the underlying data structure.
\bibliography{example_paper}

@inproceedings{feydy2019interpolating,
  title={Interpolating between optimal transport and mmd using sinkhorn divergences},
  author={Feydy, Jean and S{\'e}journ{\'e}, Thibault and Vialard, Fran{\c{c}}ois-Xavier and Amari, Shun-ichi and Trouv{\'e}, Alain and Peyr{\'e}, Gabriel},
  booktitle={The 22nd international conference on artificial intelligence and statistics},
  pages={2681--2690},
  year={2019},
  organization={PMLR}
}

@article{geng2025mean,
  title={Mean flows for one-step generative modeling},
  author={Geng, Zhengyang and Deng, Mingyang and Bai, Xingjian and Kolter, J Zico and He, Kaiming},
  journal={arXiv preprint arXiv:2505.13447},
  year={2025}
}

@article{holderrieth2024generator,
  title={Generator matching: Generative modeling with arbitrary markov processes},
  author={Holderrieth, Peter and Havasi, Marton and Yim, Jason and Shaul, Neta and Gat, Itai and Jaakkola, Tommi and Karrer, Brian and Chen, Ricky TQ and Lipman, Yaron},
  journal={arXiv preprint arXiv:2410.20587},
  year={2024}
}

@article{frans2024one,
  title={One step diffusion via shortcut models},
  author={Frans, Kevin and Hafner, Danijar and Levine, Sergey and Abbeel, Pieter},
  journal={arXiv preprint arXiv:2410.12557},
  year={2024}
}

@article{oudet2011modica,
  title={A Modica-Mortola approximation for branched transport and applications},
  author={Oudet, Edouard and Santambrogio, Filippo},
  journal={Archive for rational mechanics and analysis},
  volume={201},
  number={1},
  pages={115--142},
  year={2011},
  publisher={Springer}
}

@article{liu2022flow,
  title={Flow straight and fast: Learning to generate and transfer data with rectified flow},
  author={Liu, Xingchao and Gong, Chengyue and Liu, Qiang},
  journal={arXiv preprint arXiv:2209.03003},
  year={2022}
}

@article{ho2020denoising,
  title={Denoising diffusion probabilistic models},
  author={Ho, Jonathan and Jain, Ajay and Abbeel, Pieter},
  journal={Advances in neural information processing systems},
  volume={33},
  pages={6840--6851},
  year={2020}
}

@article{lipman2022flow,
  title={Flow matching for generative modeling},
  author={Lipman, Yaron and Chen, Ricky TQ and Ben-Hamu, Heli and Nickel, Maximilian and Le, Matt},
  journal={arXiv preprint arXiv:2210.02747},
  year={2022}
}

@article{lippmann2022theory,
  title={Theory and approximate solvers for branched optimal transport with multiple sources},
  author={Lippmann, Peter and Fita Sanmart{\'\i}n, Enrique and Hamprecht, Fred A},
  journal={Advances in Neural Information Processing Systems},
  volume={35},
  pages={267--279},
  year={2022}
}

@article{tong2023improving,
  title={Improving and generalizing flow-based generative models with minibatch optimal transport},
  author={Tong, Alexander and Fatras, Kilian and Malkin, Nikolay and Huguet, Guillaume and Zhang, Yanlei and Rector-Brooks, Jarrid and Wolf, Guy and Bengio, Yoshua},
  journal={arXiv preprint arXiv:2302.00482},
  year={2023}
}

@article{tang2025branched,
  title={Branched Schr$\backslash$" odinger Bridge Matching},
  author={Tang, Sophia and Zhang, Yinuo and Tong, Alexander and Chatterjee, Pranam},
  journal={arXiv preprint arXiv:2506.09007},
  year={2025}
}

@article{bernot2005traffic,
  title={Traffic plans},
  author={Bernot, Marc and Caselles, Vicent and Morel, Jean-Michel},
  journal={Publicacions Matem{\`a}tiques},
  pages={417--451},
  year={2005},
  publisher={JSTOR}
}

@article{maddalena2003variational,
  title={A variational model of irrigation patterns},
  author={Maddalena, Francesco and Taglialatela, Giovanni and Morel, Jean-Michel},
  journal={Interfaces and Free Boundaries},
  volume={5},
  number={4},
  pages={391--416},
  year={2003}
}

@article{xia2003optimal,
  title={Optimal paths related to transport problems},
  author={Xia, Qinglan},
  journal={Communications in Contemporary Mathematics},
  volume={5},
  number={02},
  pages={251--279},
  year={2003},
  publisher={World Scientific}
}

@article{buttazzo2003optimal,
  title={Optimal transportation networks as free Dirichlet regions for the Monge-Kantorovich problem},
  author={Buttazzo, Giuseppe and Stepanov, Eugene},
  journal={Annali della Scuola Normale Superiore di Pisa-Classe di Scienze},
  volume={2},
  number={4},
  pages={631--678},
  year={2003}
}

@article{brasco2011benamou,
  title={A Benamou--Brenier approach to branched transport},
  author={Brasco, Lorenzo and Buttazzo, Giuseppe and Santambrogio, Filippo},
  journal={SIAM journal on mathematical analysis},
  volume={43},
  number={2},
  pages={1023--1040},
  year={2011},
  publisher={SIAM}
}

@inproceedings{cuturi2013sinkhorn,
  title={Sinkhorn distances: Lightspeed computation of optimal transport},
  author={Cuturi, Marco},
  booktitle={Advances in neural information processing systems},
  pages={2292--2300},
  year={2013}
}

@article{grathwohl2018ffjord,
  title={Ffjord: Free-form continuous dynamics for scalable reversible generative models},
  author={Grathwohl, Will and Chen, Ricky TQ and Bettencourt, Jesse and Sutskever, Ilya and Duvenaud, David},
  journal={arXiv preprint arXiv:1810.01367},
  year={2018}
}

@book{villani2008optimal,
  title={Optimal transport: old and new},
  author={Villani, C{\'e}dric},
  volume={338},
  year={2008},
  publisher={Springer Science \& Business Media}
}

@article{santambrogio2015optimal,
  title={Optimal transport for applied mathematicians},
  author={Santambrogio, Filippo},
  journal={Birk{\"a}user, NY},
  volume={55},
  number={58-63},
  pages={94},
  year={2015},
  publisher={Springer}
}

@article{chen2018neural,
  title={Neural ordinary differential equations},
  author={Chen, Ricky TQ and Rubanova, Yulia and Bettencourt, Jesse and Duvenaud, David K},
  journal={Advances in neural information processing systems},
  volume={31},
  year={2018}
}

@article{mmd,
  author    = {Arthur Gretton and
               Karsten M. Borgwardt and
               Malte J. Rasch and
               Bernhard Sch{\"{o}}lkopf and
               Alexander J. Smola},
  title     = {A Kernel Two-Sample Test},
  journal   = {J. Mach. Learn. Res.},
  volume    = {13},
  pages     = {723--773},
  year      = {2012},
  url       = {http://dl.acm.org/citation.cfm?id=2188410},
}

@article{paul2015transcriptional,
  title={Transcriptional heterogeneity and lineage commitment in myeloid progenitors},
  author={Paul, Franziska and Arkin, Yaara and Giladi, Amir and Jaitin, Diego Adhemar and Kenigsberg, Ephraim and Keren-Shaul, Hadas and Winter, Deborah and Lara-Astiaso, David and Gury, Meital and Weiner, Assaf and David, Eyal and Cohen, Nadav and Lauridsen, Felicia Kathrine and Haas, Simon and Schlitzer, Andreas and Mildner, Alexander and Ginhoux, Florent and Jung, Steffen and Trumpp, Andreas and Porse, Bo Torben and Tanay, Amos and Amit, Ido},
  journal={Cell},
  volume={163},
  number={7},
  pages={1663--1677},
  year={2015},
  publisher={Elsevier}
}

@article{pan2022tedsim,
  title={Tedsim: A simulation framework for single-cell RNA sequencing data},
  author={Pan, Xiaoyu and Fan, Hongyu and Li, Wei and Shen, Haochen and Wang, Rui and Zhou, Xiaobo},
  journal={Bioinformatics},
  volume={38},
  number={8},
  pages={2208--2214},
  year={2022},
  publisher={Oxford University Press}
}

@article{liu2023generalized,
  title={Generalized Schr$\backslash$" odinger Bridge Matching},
  author={Liu, Guan-Horng and Lipman, Yaron and Nickel, Maximilian and Karrer, Brian and Theodorou, Evangelos A and Chen, Ricky TQ},
  journal={arXiv preprint arXiv:2310.02233},
  year={2023}
}
\bibliographystyle{icml2026}

\newpage
\appendix
\onecolumn
\section{Background (Extended)}
\textbf{Continuous Normalizing Flow}
\label{sec:cnf}
A Continuous Normalizing Flow (CNF) \cite{chen2018neural} is a generative model that defines a probability density path through a Neural Ordinary Differential Equation (ODE):
\begin{equation*}
  \label{eq:cnf-ode}
  \frac{d}{dt} x_t = v_\theta(x_t,t), \qquad x_{t=0} \sim \mu_0.
\end{equation*}
Let $\Phi_t$ be the flow map associated with this ODE, which transports a particle from its initial condition at time $0$ to its location at time $t$. The pushforward density $\rho_t = (\Phi_t)_\# \rho_0$ that evolves under this dynamics \emph{necessarily} satisfies the \emph{continuity equation} with the parameterized velocity field $v_\theta$. The continuity equation encodes the law of mass conservation:
\begin{align}
  \partial_t \rho_t + \nabla \cdot (\rho_t v_t) &= 0 \nonumber
  \quad \text{on } \Omega \times (0,1), \\ \nonumber
  \rho_{t=0}&=\rho_0,\quad \rho_{t=1}=\rho_1.\nonumber
\end{align}
A key result is the instantaneous change of variables formula, which describes how the log density evolves along a trajectory:
\begin{equation*}
  \label{eq:cnf-logdet}
  \frac{d}{dt}\,\log \rho_t(x_t) = -\,\nabla\!\cdot v_\theta(x_t,t). 
\end{equation*}
This allows for a likelihood calculation by integrating this quantity over time. Training can be done by directly maximizing likelihood (integrating \eqref{eq:cnf-logdet}).

\textbf{Flow Matching.} The core idea of Flow Matching (FM)\citep{lipman2022flow, tong2023improving} is to train a CNF by directly regressing its velocity field $v_\theta$ towards a target vector field $u_t$ that generates a desired probability path. The Flow Matching objective is: $\mathcal{L}_{\text{FM}}(\theta)
  = \mathbb{E}_{t\sim \mathcal{U}[0,1],\, x\sim \rho_t}\!\left[\|v_\theta(x,t)-u_t(x)\|^2\right]$.

A critical challenge is that sampling $x \sim \rho_t$ from the marginal path at arbitrary times is typically intractable. Conditional Flow Matching (CFM) \citep{lipman2022flow} provides a solution by constructing the marginal path as a mixture of simpler and tractable conditional paths. Let $z$ be a conditioning variable with distribution $q(z)$. We define the marginal path as follows: $\rho_t(x) = \int \rho_t(x\mid z)  q(z)\,dz$, where each conditional path $\rho_t(x|z)$ is generated by a corresponding conditional vector field $u_t(x\mid z)$. The marginal field $u_t(x)$ that generates $\rho_t$ is then given by:
\begin{equation*}
  \label{eq:marginal-field}
  u_t(x) \;=\; \mathbb{E}_{z\sim q(z\mid x)}\!\left[u_t(x\mid z)\right] = \mathbb{E}_{z\sim q}\!\left[ \frac{\rho_t(x\mid z)}{\rho_t(x)}\,u_t(x\mid z) \right],
\end{equation*}
where $q(z\mid x)$ is the posterior. The key theorem is to minimize the following \emph{Conditional Flow Matching} objective:
\begin{equation}
  \label{eq:cfm}
  \mathcal{L}_{\text{CFM}}(\theta)
  = \mathbb{E}_{t\sim\mathcal{U}[0,1],\, z\sim q,\, x\sim \rho_t(\cdot\mid z)}\!\left[\|v_\theta(x,t)-u_t(x\mid z)\|^2\right]
\end{equation}
yields the same gradient for $\theta$ as minimizing the intractable $\mathcal{L}_{\text{FM}}(\theta)$. This makes CFM a practical objective, as it only requires sampling from conditional paths $\rho_t(x|z)$ and knowing their closed-form drifts $u_t(x|z)$.

The flexibility of CFM lies in the choice of conditional paths. The coupling $q(z)$ is the independent joint distribution $q(x_0)q(x_1)$, so $z=(x_0, x_1)$. A common conditional path is a Gaussian bridge: $\rho_t(x\mid z) = \mathcal{N}(x \mid \mu_t, \sigma^2 I)$, where $\mu_t = (1-t)x_0 + t x_1$ is a linear interpolation. The simple constant drift that generates this path is $u_t(x\mid z)=x_1-x_0$. The optimal transport CFM (OT-CFM) \citep{tong2023improving} method uses an optimal coupling to define conditionals. Here, $z=(x_0, x_1)$ is sampled from an OT plan $\pi$ between $\mu_0$ and $\mu_1$ \ref{sec:static_ot}. 
In practice, the OT plan $\pi$ is efficiently approximated using mini-batch OT, which has been shown to work well empirically.

\section{Modica-Mortola Branched Flows}
\label{appendix:MM}
\begin{figure*}[h!]
    \centering
    \includegraphics[width=0.99\linewidth]{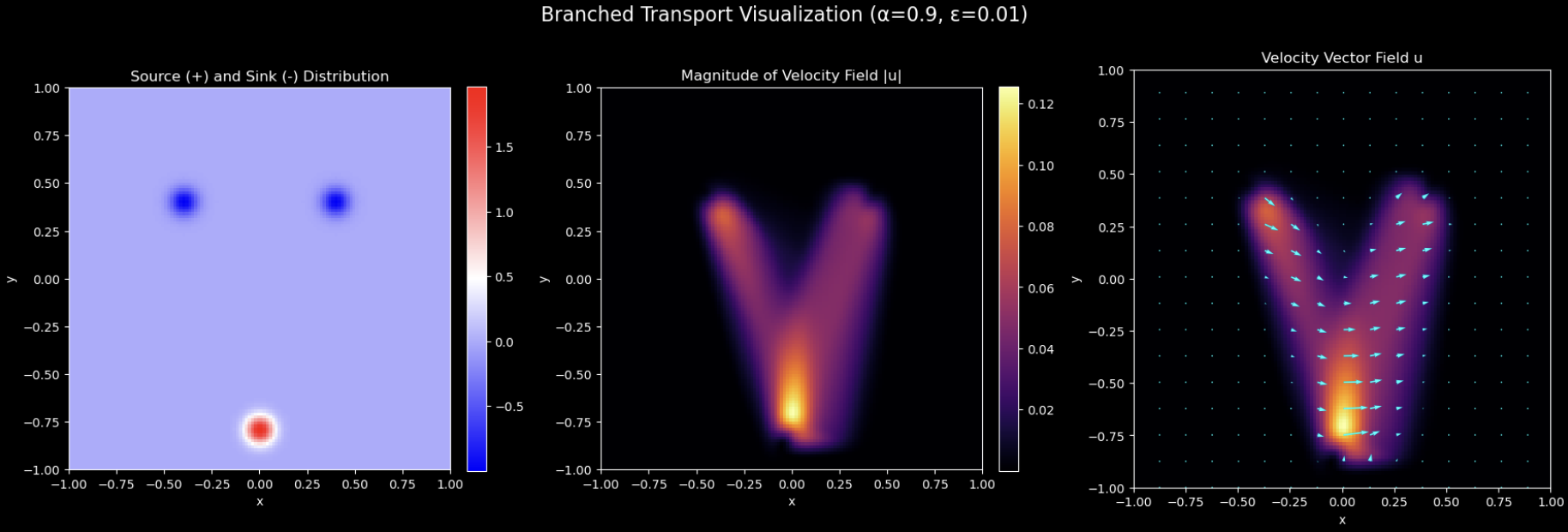}
    \includegraphics[width=0.99\linewidth]{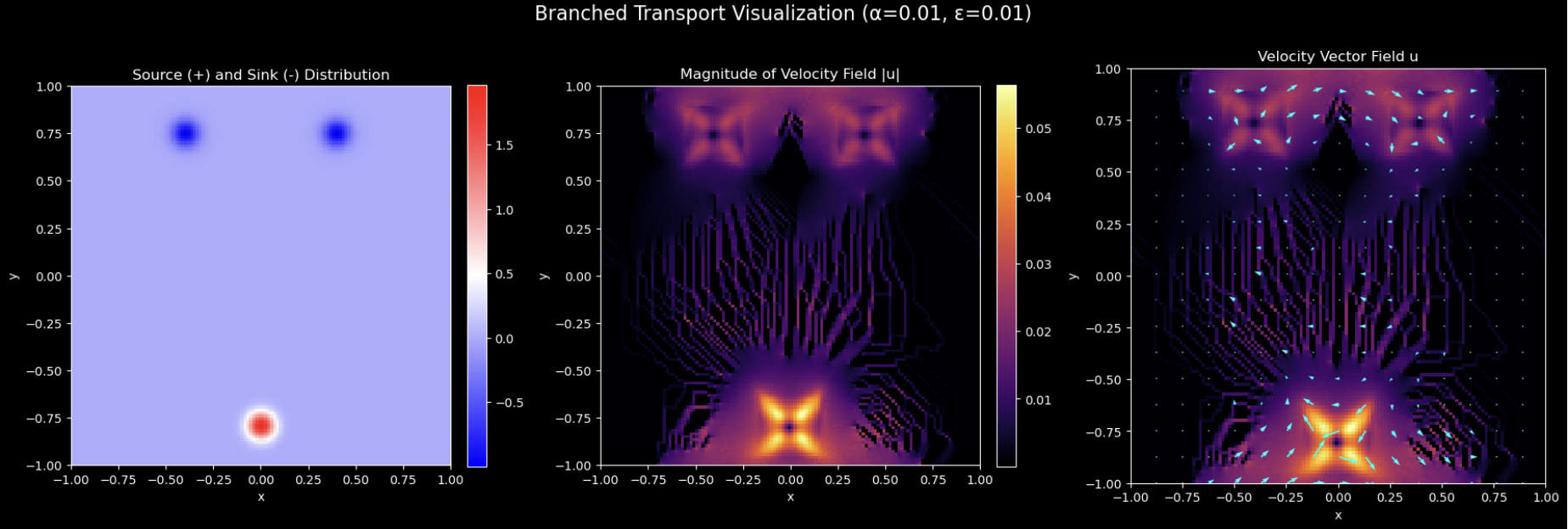}
    \caption{Modica-Mortola solver results using a neural network-based field parametrization. The difference between the top and bottom rows illustrates the sensitivity of the method to optimization parameters, specifically the learning rate. We observe that the method is highly sensitive to initialization and optimization settings.}
    \label{fig:mm_exp}
\end{figure*}

\cite{oudet2011modica} proposed a theoretically grounded approximation inspired by the Modica-Mortola framework used in phase-field modeling. This framework replaces the singular branched transport problem with a sequence of regularized elliptic energy functionals $M_\lambda^{\alpha}$, defined over the more regular space of vector fields $\mathcal{H}^1$. For a given vector field $F(x)$, the approximating functional is defined as:
\begin{equation}
    M_\lambda^{\alpha}(F) = \lambda^{\gamma_1} \int_{\Omega} |F(x)|^{\alpha} dx + \lambda^{\gamma_2} \int_{\Omega} |\nabla F(x)|^2 dx,
    \label{eq:modica_mortola_approx_appendix}
\end{equation}
where $\lambda > 0$ is a small regularization parameter. The exponents $\gamma_1$ and $\gamma_2$ are derived from the transport dimension $d$ and the cost exponent $\alpha$ to ensure correct scaling behavior. As described by \cite{oudet2011modica}, the functional in ~\eqref{eq:modica_mortola_approx_appendix} consists of two competing terms:
\begin{enumerate}
    \item \textbf{A concave potential term} ($\int |F|^{\alpha} dx$): This term acts analogously to a double-well potential, encouraging the magnitude of the vector field to be either zero or arbitrarily large, thereby promoting the formation of sparse high-intensity pathways.
    \item \textbf{A Dirichlet energy term} ($\int |\nabla F|^2 dx$): This Sobolev penalty penalizes spatial variations, enforcing smoothness and regularity on the vector field.
\end{enumerate}

The central theoretical guaranty is that as $\lambda \to 0$, the functionals $M_\lambda^{\alpha}$ $\Gamma$-converge to the original branched transport energy $M^{\alpha}$. This ensures that minimizers of the regularized problem converge to a minimizer of the intractable branched transport problem.

Although the original implementation by \cite{oudet2011modica} relied on a staggered grid discretization and FFT-based projections to enforce the divergence constraint, we reformulated this variational principle within a deep learning framework.

In our experiments, we parameterized the vector field $F$ as a coordinate-based neural network (an MLP) $F_\theta: \Omega \to \mathbb{R}^d$, implemented in PyTorch. Unlike grid-based methods, this allows for a mesh-free representation of the transport density. The optimization objective $\mathcal{L}(\theta)$ is composed of the relaxed energy functional and a soft penalty for the divergence constraint:
\begin{equation}
    \mathcal{L}(\theta) = M_\lambda^{\alpha}(F_\theta) + \beta \| \nabla \cdot F_\theta - (\mu_0 - \mu_1) \|^2,
\end{equation}
where $\beta$ is a penalty weight (In practice, we find it important to put this value high, in our experiments, the method worked only with $\beta>10k$). We utilized PyTorch's automatic differentiation (\texttt{torch.autograd}) to compute the spatial Jacobian $\nabla F_\theta$ for the Dirichlet term and the divergence $\nabla \cdot F_\theta$ for the constraint term exact to the network precision. The integrals were approximated by Monte Carlo sampling in the domain $\Omega$.

We evaluated this approach in a simple setting with a single source and two symmetric targets. Although the method successfully generated the branching structures shown in Figure \ref{fig:mm_exp}, we found that the training dynamics is notoriously difficult to stabilize. The optimization landscape is highly sensitive to hyperparameters; for instance, training diverged significantly at a learning rate of $1\text{e-}4$, yet converged to a reasonable solution at $1\text{e-}5$.

We attribute this difficulty to the inherent nature of the flux variable in the Modica-Mortola formulation. Since $F$ represents a mass flux (density $\times$ velocity), optimal solutions tend toward singularity (Dirac structures) as $\lambda \to 0$. Neural networks often struggle with such high-frequency spectral components (spectral bias), making the simultaneous minimization of the concave potential and the divergence constraint numerically unstable without careful scheduling of the regularization parameter $\lambda$.

\section{Proofs}
\label{sec:proofs}
\section{Formalization of Branching Intuition}
\label{sec:intuition_proofs}

We consider a dynamic problem on a domain $\Omega \subset \mathbb{R}^d$ over a time interval $t \in [0,1]$. The goal is to transport a source distribution $\rho_0$ to a target $\rho_1$ while minimizing a functional composite energy. The density $\rho_t$ evolves according to the continuity equation $\partial_t \rho + \nabla \cdot (\rho v) = 0$. The total energy functional is given by:

\begin{equation}
    J(\rho,v) = \underbrace{\int_0^1 \int_\Omega \rho_t(x) \|v_t(x)\|^\alpha \, dx \, dt}_{\mathcal{T}(\rho,v)} + \lambda \underbrace{\int_0^1 \int_\Omega \rho_t(x) \|\nabla v_t(x)\|_F^2 \, dx \, dt}_{\mathcal{C}(\rho,v)}
    \label{eq:total_energy}
\end{equation}

where $\alpha \in (0,1)$ is the sub-linear transport exponent, $\lambda > 0$ is the cohesion weight, and $\|\cdot\|_F$ denotes the Frobenius norm.

\subsection*{Symmetric Toy Problem Setup}
To analyze the branching behavior, we consider a simplified symmetric case:
\begin{itemize}
    \item \textbf{Source:} A mass distribution concentrated at $P=(0,0)$ at $t=0$.
    \item \textbf{Targets:} Two symmetric targets $Q_1=(-w,h)$ and $Q_2=(w,h)$ in $t=1$.
    \item \textbf{Particle Model:} We model the mass as a connected cloud of characteristic width $\epsilon$.
    \item \textbf{Trajectory Strategy:} We parameterize the path by a branching time $\tau \in [0,1]$.
    \begin{itemize}
        \item For $t \in [0, \tau)$, the mass moves as a single cluster (Trunk).
        \item For $t \in [\tau, 1]$, the mass splits and moves towards the respective targets (Branches).
    \end{itemize}
\end{itemize}

The optimization problem reduces to finding $\tau$ that minimizes the trade-off between the \textbf{Transport Cost} (path length/speed efficiency) and the \textbf{Cohesion Cost} (velocity field regularity).

\bigskip

\begin{proposition}[Zero-Cost of Uniform Motion]
\label{prop:zero_cost}
If the velocity field $v(x,t)$ represents a spatially uniform translation (rigid body motion without rotation) on the support of $\rho_t$, the Cohesion Cost density is zero.
\end{proposition}

\begin{proof}
Let $\text{supp}(\rho_t)$ denote the spatial support of the density at time $t$. Assume $v(x,t) = \mathbf{u}(t)$ for all $x \in \text{supp}(\rho_t)$, where $\mathbf{u}(t)$ depends only on time.
The Frobenius norm of the velocity gradient is defined as:
\begin{equation}
    \|\nabla v\|_F^2 = \sum_{i,j} \left( \frac{\partial v_i}{\partial x_j} \right)^2
\end{equation}
Since $v$ is constant with respect to the spatial coordinates $x$, the partial derivatives $\frac{\partial v_i}{\partial x_j}$ vanish identically. Consequently:
\begin{equation}
    \int_{\text{supp}(\rho_t)} \rho_t(x) \|\nabla v(x,t)\|_F^2 \, dx = \int \rho_t(x) \cdot 0 \, dx = 0
\end{equation}
Thus, the \emph{Trunk} phase of the trajectory incurs zero penalty from the cohesion term.
\end{proof}

\bigskip

\begin{proposition}[The Cost of Separation]
\label{prop:spatial_conflict}
If a continuous velocity field separates the mass into distinct directions over a transition region of width $\epsilon$, the Cohesion Cost scales as $\mathcal{O}(1/\epsilon)$.
\end{proposition}

\begin{proof}
Consider a 1D cross-section of the particle cloud at the moment of splitting, defined on the interval $x \in [-\epsilon, \epsilon]$. To achieve separation:
\begin{itemize}
    \item For $x \in [-\epsilon, 0)$, $v(x) \approx -u$ (moving left).
    \item For $x \in (0, \epsilon]$, $v(x) \approx +u$ (moving right).
\end{itemize}
Assuming $v$ is differentiable, it must transition from $-u$ to $+u$ over a distance of $2\epsilon$. By the Mean Value Theorem applied to each component, there exists a point $x_0$ in the interval where the gradient magnitude satisfies $|\nabla v(x_0)| \geq \frac{2u}{2\epsilon} = \frac{u}{\epsilon}$.
The contribution to the energy integral approximates: 
\begin{equation}
    \int_{-\epsilon}^{\epsilon} \rho \|\nabla v\|^2 dx \approx \rho_{\text{avg}} \cdot (2\epsilon) \cdot \left( \frac{u}{\epsilon} \right)^2 \propto \frac{1}{\epsilon}
\end{equation}
As $\epsilon \to 0$, this cost diverges, creating an infinite barrier against the immediate separation of a singular measure.
\end{proof}

\bigskip

\begin{proposition}[Time-Compression]
\label{prop:time_compression}
For $\alpha \in (0,1)$, the transport cost of traveling a fixed distance $D$ decreases as the duration of the travel $T$ decreases.
\end{proposition}

\begin{proof}
Consider a particle traveling a distance $D$ over duration $T$ with constant speed $|v| = D/T$. The Transport Cost is
\begin{equation}
    C(T) = \int_0^T |v|^\alpha dt = T \cdot \left( \frac{D}{T} \right)^\alpha = D^\alpha T^{1-\alpha}
\end{equation}
We analyze the sensitivity of the cost to the duration $T$:
\begin{equation}
    \frac{d C}{dT} = D^\alpha (1-\alpha) T^{-\alpha}
\end{equation}
Since $\alpha \in (0,1)$, the term $(1-\alpha)$ is strictly positive. Therefore, $\frac{dC}{dT} > 0$.
This implies that reducing travel time $T$ (and consequently increasing velocity) strictly reduces the total energy cost. This property favors ``impulsive'' motion: waiting in the trunk (where cost is low) and then bursting to the target in the remaining short time window.
\end{proof}

\bigskip

\begin{lemma}[Existence of Optimal Branching Time]
For a sufficiently large cohesion weight $\lambda$ and $\alpha \in (0,1)$, the optimal branching time $\tau^*$ satisfies $\tau^* > 0$. This implies that a Y-shaped trajectory is energetically superior to a V-shaped trajectory (where $\tau=0$).
\end{lemma}

\begin{proof}
We minimize the total energy $E(\tau)$ with respect to the branching time $\tau \in [0, 1)$. Let the trunk length be $\tau h$ and the remaining branch length be $L_B(\tau) = \sqrt{w^2 + h^2(1-\tau)^2}$.

The total energy is the sum of the trunk transport, branch transport, and branch cohesion costs:
\begin{equation}
    E(\tau) = \underbrace{(\tau h)^\alpha \tau^{1-\alpha}}_{\text{Trunk } \mathcal{T}} + \underbrace{L_B(\tau)^\alpha (1-\tau)^{1-\alpha}}_{\text{Branch } \mathcal{T}} + \underbrace{\lambda K_{\text{split}} (1-\tau)}_{\text{Branch } \mathcal{C}}
\end{equation}
Simplifying the terms (using $\tau^\alpha \tau^{1-\alpha} = \tau$):
\begin{equation}
    E(\tau) = h^\alpha \tau + \left( \sqrt{w^2 + h^2(1-\tau)^2} \right)^\alpha (1-\tau)^{1-\alpha} + \lambda K_{\text{split}} (1-\tau)
\end{equation}
We evaluate the derivative at $\tau=0$ (the V-shaped configuration) to check for a descent direction.
\begin{equation}
    E'(0) = h^\alpha + \left[ \frac{d}{d\tau} \left( L_B(\tau)^\alpha (1-\tau)^{1-\alpha} \right) \right]_{\tau=0} - \lambda K_{\text{split}}
\end{equation}
The dominant term in the bracket comes from differentiating $(1-\tau)^{1-\alpha}$. At $\tau=0$, this derivative is $-(1-\alpha)$. The cohesion term contributes $-\lambda K_{\text{split}}$.
Thus, the derivative has the form:
\begin{equation}
    E'(0) \approx h^\alpha - (1-\alpha)L_B(0)^\alpha - \lambda K_{\text{split}}
\end{equation}
Since $L_B(0) > h$ (the hypotenuse is longer than the vertical leg) and $\lambda K_{\text{split}} > 0$, for sufficiently large $\lambda$ or small $\alpha$, we have:
\begin{equation}
    E'(0) < 0
\end{equation}
Since the derivative is negative at $\tau=0$, the energy decreases as $\tau$ increases. Therefore, the minimum must occur at some $\tau^* > 0$, confirming the optimality of the Y-shape.
\end{proof}
\section{Discussion and Limitations}
\paragraph{Well-posedness for $\alpha\in(0,1)$.}
For $\alpha\in(0,1)$, the map $v\mapsto \|v\|^\alpha$ is sublinear, which induces the time-compression effect, but can also cause the continuous-time problem to degenerate if arbitrarily large speeds are allowed. A standard remedy is to enforce at least one of the following:
\begin{itemize}
    \item a hard speed bound $\|v(t,x)\|\le v_{\max}$ almost everywhere, or
    \item a small convex stabilizer, e.g.\ $+\mu\int_0^1\!\!\int_\Omega \rho\,\|v\|^2\,dx\,dt$ with $\mu>0$, or
    \item a smooth surrogate $\big(\|v\|^2+\delta^2\big)^{\alpha/2}$ with $\delta>0$ This can be useful for stable gradients near $v=0$.
\end{itemize}
Any of these prevents ``infinite-speed, zero-time'' collapse while preserving the qualitative preference for bursty motion. \textbf{But in our experiments, we did not face these challenges, so we avoided these regularizations.}


\paragraph{Density-weighting caveat.}
Because $\mathcal{C}$ is weighted by $\rho$, sharp gradients in $v$ are cheap, where $\rho$ is close to zero. If the model is free to create low-density \emph{gaps} in the transition layer, it can partially bypass the separation barrier. In practice, one can mitigate this with density regularization (e.g.\ entropy) or by constraining how $\rho$ is represented.

\paragraph{Deterministic ODE caveat (literal splitting of a point mass).}
A deterministic flow map sends a Dirac mass to a Dirac mass; it cannot turn one particle into two. In particle implementations, ``branching'' should therefore be interpreted as \emph{different particles} (initially distributed under $\rho_0$) peeling into distinct modes, rather than a literal split of a single trajectory.

In general, it is important to note that if $\alpha\ge 1$, the time-compression advantage disappears (the speed penalty is no longer sublinear), weakening the incentive for delayed bursting. 

\section{Additional Experiments and Illustrations}
\subsection{LiDAR}
\label{data:lidar}
Airborne LiDAR \cite{liu2023generalized} terrain tile with ground and low-vegetation; each sample is a 3D point $(x,y,z)$ with optional attributes (intensity, return number, total returns, scan angle, class). After filtering non-ground classes, removing outliers with a radius/k-NN check, and Poisson-disk (or voxel-grid) subsampling, we retain ~5k points that preserve ridges, slopes, and basins. Coordinates are centered and scaled to a unit box (optionally PCA-aligned). We estimate per-point normals via local PCA and build a k-NN surface graph $(k \in [8,16])$ with Euclidean edge weights as a geodesic proxy; during optimization, states are projected back to the nearest tangent patch to stay on-manifold. A single source distribution sits on a lower slope; four target clouds are disjoint regions on ridges/basins (indices provided for reproducibility).

\subsection{Image Data}
\begin{figure}[h]
    \centering
    \includegraphics[width=\linewidth]{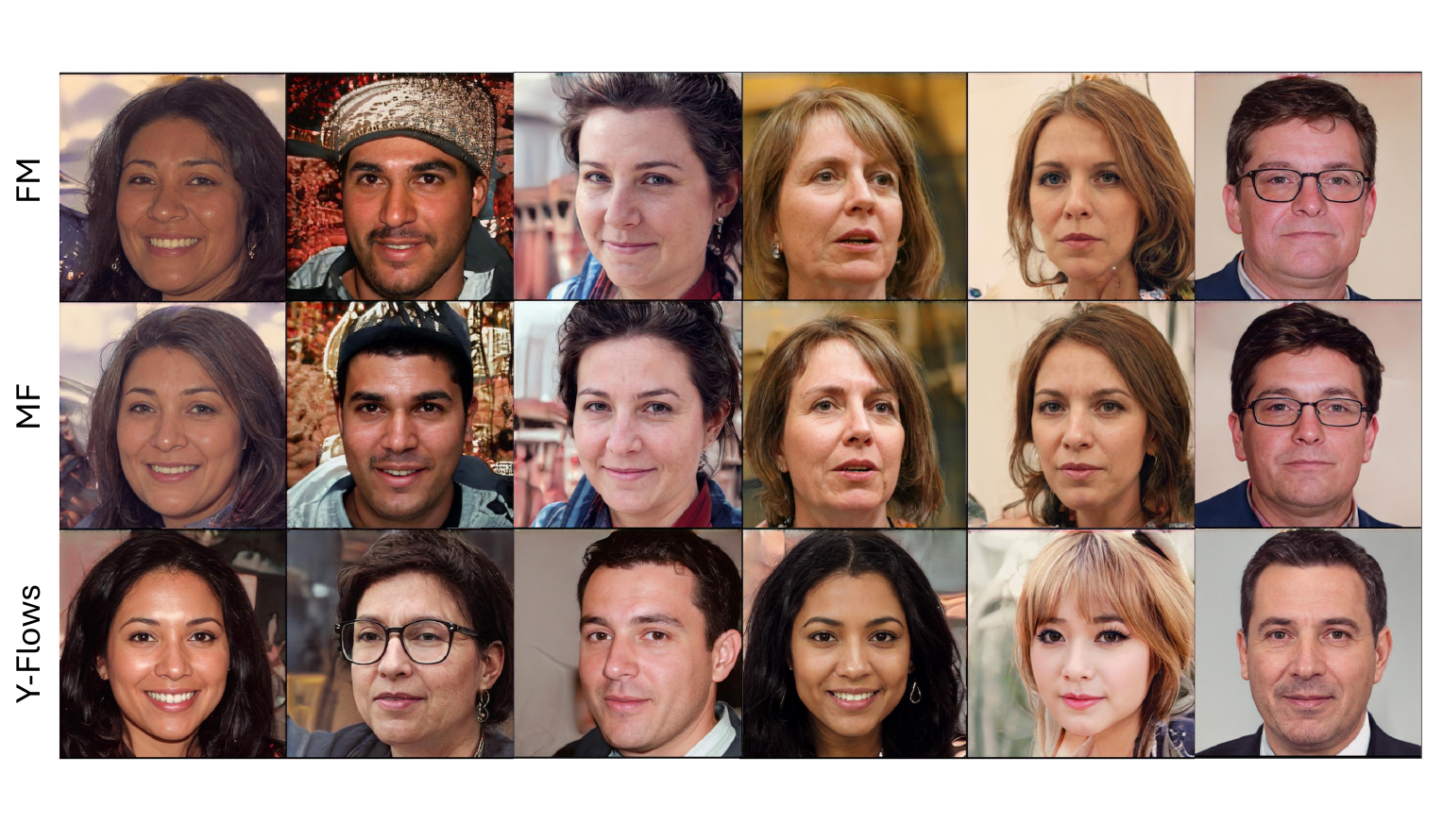}
    \caption{Image generation. Each row depicts a result of a generative model predicting a sample from the target distribution $\mu_1$. Samples from the source distribution $\mu_0$ are fixed and are the  same for all models.}
    \label{fig:generation}
\end{figure}

The implementation of \emph{Flow Matching} was based on: \newline
\href{https://github.com/facebookresearch/flow_matching.git}{\texttt{facebookresearch/flow\_matching}}.

The implementation \emph{Mean Flows} was based on: \newline
\href{https://github.com/Gsunshine/py-meanflow.git}{\texttt{Gsunshine/py-meanflow}}.

The experiment on \emph{Domain Translation} was inspired by: \newline
\href{https://github.com/milenagazdieva/LightUnbalancedOptimalTransport.git}{\texttt{milenagazdieva/LightUnbalancedOptimalTransport}}.

FFHQ \emph{encoded dataset and decoder} are taken from: \newline 
\href{https://github.com/podgorskiy/ALAE/tree/master}{\texttt{podgorskiy/
ALAE}}

\subsection{Use of LLM}
We used an LLM for grammar editing.

\end{document}